\documentclass[sigconf]{acmart} % camera-ready
\AtBeginDocument{%
  }

\acmConference[ArXiv '26]{Make sure to enter the correct
  conference title from your rights confirmation email}{February 09,
  2026}{ArXiv}
\usepackage{algorithm}
\usepackage{algorithmic}
\usepackage{enumitem}
\usepackage{float}
\usepackage{caption}
\usepackage{cuted}
\usepackage{listings}
\usepackage{tcolorbox}
\tcbuselibrary{listings,breakable}
\usepackage{mdframed}
\usepackage{seqsplit}
\begin{document}

%%
%% The "title" command has an optional parameter,
%% allowing the author to define a "short title" to be used in page headers.
\title{FM SO.P: A Progressive Task Mixture Framework with Automatic Evaluation for Cross-Domain SOP Understanding}

%%
%% The "author" command and its associated commands are used to define
%% the authors and their affiliations.
%% Of note is the shared affiliation of the first two authors, and the
%% "authornote" and "authornotemark" commands
%% used to denote shared contribution to the research.
\author{Siyuan Huang$^{12}$, Ziyu Wang$^1$, Chao Pan$^1$, Han Zhao$^{13}$}
% \authornote{Both authors contributed equally to this research.}
% \orcid{1234-5678-9012}
% \author{G.K.M. Tobin}
% \authornotemark[1]
% \email{webmaster@marysville-ohio.com}
\affiliation{%
  \institution{$^{1}$Amazon, $^{2}$Johns Hopkins University, $^{3}$University of Illinois Urbana-Champaign}
  % \city{Dublin}
  % \state{Ohio}
  \country{}
}
\email{shuan124@jhu.edu, ziywan@amazon.com, pchao@amazon.com, hanzhao@illinois.edu}

% \author{Lars Th{\o}rv{\"a}ld}
% \affiliation{%
%   \institution{The Th{\o}rv{\"a}ld Group}
%   \city{Hekla}
%   \country{Iceland}}
% \email{larst@affiliation.org}

% \author{Valerie B\'eranger}
% \affiliation{%
%   \institution{Inria Paris-Rocquencourt}
%   \city{Rocquencourt}
%   \country{France}
% }

% \author{Aparna Patel}
% \affiliation{%
%  \institution{Rajiv Gandhi University}
%  \city{Doimukh}
%  \state{Arunachal Pradesh}
%  \country{India}}

% \author{Huifen Chan}
% \affiliation{%
%   \institution{Tsinghua University}
%   \city{Haidian Qu}
%   \state{Beijing Shi}
%   \country{China}}

% \author{Charles Palmer}
% \affiliation{%
%   \institution{Palmer Research Laboratories}
%   \city{San Antonio}
%   \state{Texas}
%   \country{USA}}
% \email{cpalmer@prl.com}

% \author{John Smith}
% \affiliation{%
%   \institution{The Th{\o}rv{\"a}ld Group}
%   \city{Hekla}
%   \country{Iceland}}
% \email{jsmith@affiliation.org}

% \author{Julius P. Kumquat}
% \affiliation{%
%   \institution{The Kumquat Consortium}
%   \city{New York}
%   \country{USA}}
% \email{jpkumquat@consortium.net}

%%
%% By default, the full list of authors will be used in the page
%% headers. Often, this list is too long, and will overlap
%% other information printed in the page headers. This command allows
%% the author to define a more concise list
%% of authors' names for this purpose.
% \renewcommand{\shortauthors}{Trovato et al.}
\newcommand{\han}[1]{\textbf{\textcolor{blue}{[Han: #1]}}}

%%
%% The abstract is a short summary of the work to be presented in the
%% article.
\begin{abstract}
Standard Operating Procedures (SOPs) are critical for enterprise operations, yet existing language models struggle with SOP understanding and cross-domain generalization. Current methods fail because joint training cannot differentiate between reasoning capabilities that SOP requires: terminology precision, sequential ordering, and constraint reasoning. We propose FM SO.P, solving these challenges through two novelties. First, we introduce progressive task mixtures that build capabilities by stages across three task types with cumulative data: concept disambiguation for terminology precision, action sequence understanding for procedural correctness, and scenario-aware graph reasoning for conditional logic. Second, we propose an automatic multi-agent evaluation system consisting of three agents that adaptively generate rubrics, stratified test sets, and rubric scoring, adapting to domains (e.g., temporal constraints for DMV, regulatory compliance for banking). Evaluated on SOPBench across seven domains (Bank, DMV, Healthcare, Market, University, Library, Hotel), FM SO.P achieves 48.3\% pass rate with our 32B model and 34.3\% with our opensource 7B model, matching Qwen-2.5-72B-Instruct baseline (34.4\%) with 10× fewer parameters.

\end{abstract}

\begin{teaserfigure}
  \centering
  \includegraphics[width=0.85\textwidth]{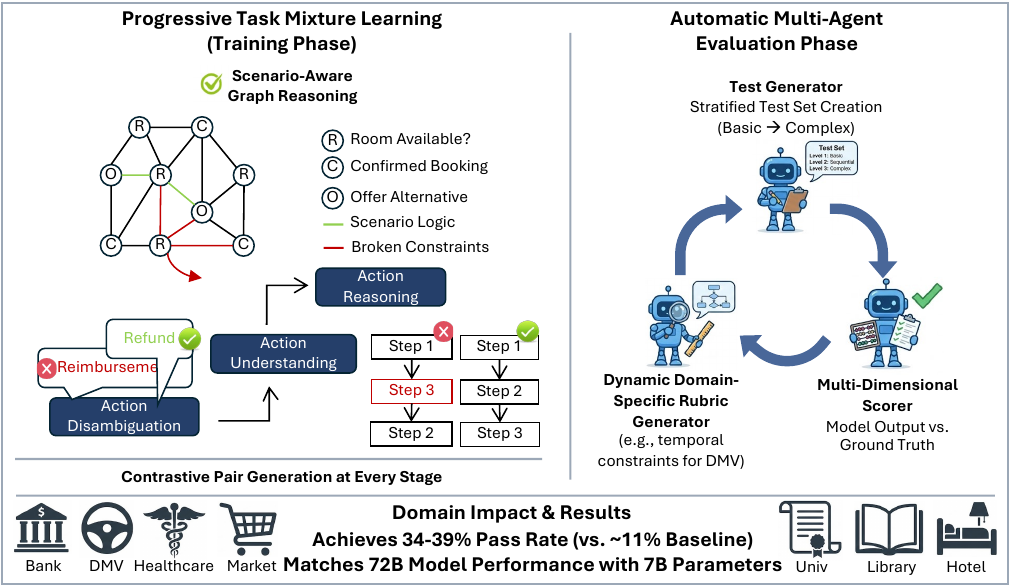}
  \vspace{-1em}
  \caption{Overview of the FM SO.P Framework. Our framework consists of two main components: (Left) Progressive Task Mixture (Training Phase) with three stages: concept disambiguation differentiating domain terminology (e.g., "refund" vs. "reimbursement"), action sequence understanding with procedural error injection, and scenario-aware graph reasoning capturing conditional dependencies. Contrastive pairs are generated at every stage with cumulative data. (Right) Automatic Multi-Agent Evaluation Phase with three agents: dynamic rubric generator creating domain-specific evaluation criteria (e.g., temporal constraints for DMV), test generator producing stratified test sets from basic to complex, and rubric scorer comparing model outputs against ground truth.}
  \label{fig:teaser}
\end{teaserfigure}

% \received{20 February 2007}
% \received[revised]{12 March 2009}
% \received[accepted]{5 June 2009}

%%
%% This command processes the author and affiliation and title
%% information and builds the first part of the formatted document.
\maketitle

% \begin{figure*}
%     \centering
%     \includegraphics[width=\linewidth]{kdd-26-teaser.pdf}
%     \caption{\textbf{Overview of the FM SO.P Framework.} \han{I'd suggest changing the title in the left panel from ``progressive extended curriculum learning'' to match the current term ``progressive task mixtures''} Our framework consists of three main components: (Left) \textbf{Progressive Task Mixture (Training Phase)} with three stages---concept disambiguation distinguishing domain terminology (e.g., ``refund'' vs. ``reimbursement''), action sequence understanding with procedural error injection, and scenario-aware graph reasoning capturing conditional dependencies. Contrastive positive-negative pairs are generated at every stage, with cumulative data composition. (Center) The trained \textbf{FM SO.P Model} achieves cross-domain generalization and deep procedural understanding. (Right) \textbf{Automatic Multi-Agent Evaluation Phase} with three agents: dynamic rubric generator creating domain-specific evaluation criteria (e.g., temporal constraints for DMV), test generator producing stratified test sets from basic to complex, and multi-dimensional scorer comparing model outputs against ground truth.}
%     \label{fig:teaser}
% \end{figure*}

% \vspace{-11.5pt}

\section{Introduction}

Standard Operating Procedures (SOPs) encode critical knowledge about policies, workflows, and decision making across diverse domains \cite{garg2025generating,grohs2023large,kulkarni2025agent,nandi2025sop,wang2025sop}. However, their complexity creates significant challenges: multi-step workflows with conditional logic \cite{nandi2025sop,xiao2024flowbench,yao2024tau}, cross-procedural dependencies \cite{brahman2023plasma,grohs2023large}, and context-sensitive decisions \cite{diao2025guidebench,wang2025sop}.

Recent progress in large language models shows impressive instruction following capabilities \cite{chung2024scaling,ouyang2022training,wang2023self,weifinetuned}, yet SOP understanding reveals unique challenges. General models often fail at procedural reasoning tasks \cite{brahman2024plasma, ruisprocedural,shen2024taskbench}, suggesting actions in incorrect order, missing critical validation steps, or inappropriately applying rules across contexts. Current training paradigms encourage shortcut pattern matching rather than deep procedural understanding \cite{du2023shortcut,tang2023large}, where models memorize isolated rules but fail to synthesize them into coherent workflows \cite{sakai2025revisiting,wanggeneralization}.

% Current methods suffer from three critical limitations. First, joint training on all task types creates conflicting optimization \cite{chai2023improving,yu2020gradient,liu2021conflict,standley2020tasks}. Procedural understanding requires different cognitive capabilities that build hierarchically \cite{dalvi2018tracking,song2023ego4d}: concept recognition (differentiating "refund" from "reimbursement"), sequential reasoning (verifying correct action ordering), and contextual constraints (evaluating temporal and state conditions). Training these jointly means models receive complex reasoning before mastering basic terminology, leading to interference. Second, fixed evaluation metrics cannot capture domain-specific requirements \cite{mathur2020tangled,sulem2018bleu,kim2023prometheus}. Banking SOPs require regulatory compliance assessment; healthcare SOPs need patient safety verification; DMV SOPs need temporal constraint validation. Fixed metrics apply uniform criteria across all domains, fundamentally failing to capture these domain-specific requirements. Third, larger models have deployment challenges. While 70B+ models show stronger performance, real-world deployment requires efficiency \cite{chitty2024llm,kwon2023efficient}: 7B models require approximately 14GB memory enabling single-GPU inference, whereas 70B+ models need 140GB+ memory, and multi-GPU parallelism.
Specifically, there are three limitations in current methods. First, joint training on all task types creates conflicting optimization \cite{chai2023improving,yu2020gradient,liu2021conflict,standley2020tasks}. Procedural understanding requires different cognitive capabilities that build hierarchically \cite{dalvi2018tracking,song2023ego4d}: concept recognition (differentiating "refund" from "reimbursement"), sequential reasoning (verifying correct action ordering), and contextual constraints (evaluating temporal and state conditions). Training these jointly means models receive complex reasoning before mastering basic terminology, leading to interference. Second, fixed evaluation metrics cannot capture domain-specific requirements \cite{mathur2020tangled,sulem2018bleu,kim2023prometheus}. Banking SOPs require regulatory compliance assessment; healthcare SOPs need patient safety verification; DMV SOPs need temporal constraint validation. Fixed metrics apply uniform criteria across all domains, fundamentally failing to capture these domain-specific requirements. Third, larger models have deployment challenges. While 70B+ models show stronger performance, real-world deployment requires efficiency \cite{chitty2024llm,kwon2023efficient}: 7B models require approximately 14GB memory enabling single-GPU inference, whereas 70B+ models need 140GB+ memory, and multi-GPU parallelism.

We propose FM SO.P, solving these limitations through two novelties. First, we introduce \textbf{progressive task mixtures} that systematically build capabilities through three stages with cumulative data: (1) concept disambiguation establishing domain terminology precision, (2) action sequence understanding building sequential reasoning, and (3) scenario-aware graph reasoning enabling conditional logic. Unlike curriculum learning \cite{bengio2009curriculum,soviany2022curriculum,wang2021survey} ordering samples by difficulty, our framework structures learning across different task types, with each stage retaining all previous data to eliminate conflicting optimization.

Second, we propose an \textbf{automatic multi-agent evaluation system} that provides adaptive evaluation at scale. Agent 1 analyzes domain characteristics to generate adaptive rubrics (e.g., "temporal constraint validation" for DMV, "regulatory compliance" for banking). Agent 2 creates stratified test sets across complexity levels. Agent 3 performs rubric scoring. This enables evaluation that automatically adapts to different domains.

Experiments across seven domains demonstrate large improvements. FM SO.P (7B) achieves 34.33\% pass rate, matching Qwen-2.5-72B-Instruct (34.44\%) with 10$\times$ fewer parameters. FM SO.P (32B) reaches 48.30\%, exceeding the 72B baseline by +13.86\% while using 2.25$\times$ fewer parameters.

To summarize, our contributions include: (1) \textbf{Progressive task mixtures} that build SOP understanding through staged training across different tasks with cumulative data; (2) \textbf{Automatic multi-agent evaluation} adaptively generates rubrics, enabling scalable evaluation that adapts to different domains; and (3) Demonstration that 7B models with our method match 72B baseline performance, providing practical deployment.

\begin{figure*}[!t]
    \centering
    \includegraphics[width=0.85\linewidth]{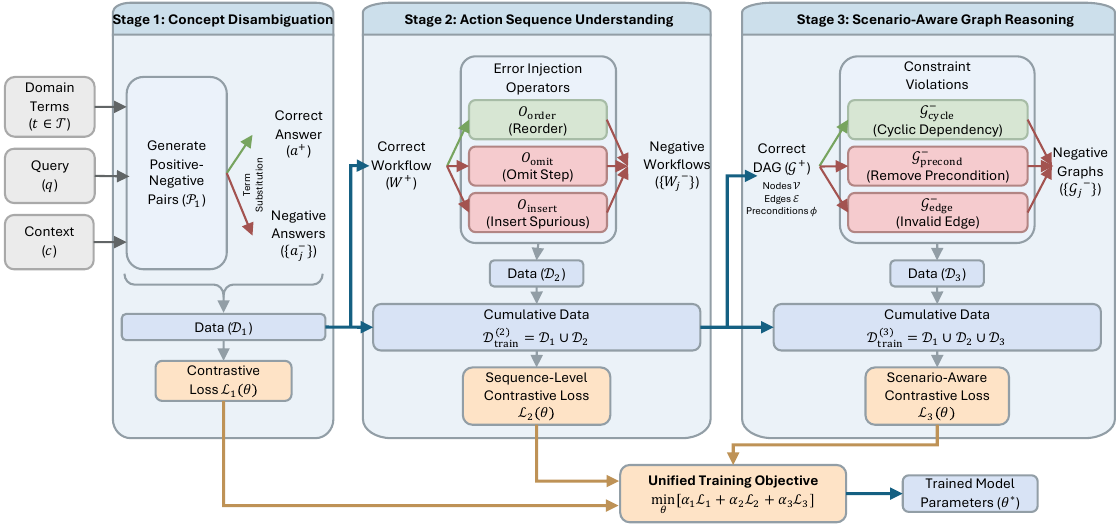}
    \caption{Progressive Task Mixture Architecture. Our three-stage framework builds SOP understanding through cumulative data. Stage 1 focuses on concept disambiguation using term substitution. Stage 2 adds action sequence understanding through procedural error injection (reordering, omission, insertion). Stage 3 incorporates scenario-aware graph reasoning with constraint violations (cycles, preconditions, invalid edges). Each stage retains all previous data ($\mathcal{D}^{(k)}_{\text{train}} = \bigcup_{i=1}^{k} \mathcal{D}_i$) and optimizes stage-specific contrastive losses, finally represented by $\min_{\theta}[\alpha_1\mathcal{L}_1 + \alpha_2\mathcal{L}_2 + \alpha_3\mathcal{L}_3]$.}
    \label{fig:curriculum_architecture}
\end{figure*}

\section{Method}

\subsection{Problem Formulation and Motivation}

\subsubsection{Formal Problem Definition}
Let $\mathcal{D} = \{d_1, d_2, \ldots, d_N\}$ denote a collection of domain-specific SOPs, where each SOP $d_i$ consists of a structured document containing procedural knowledge. An SOP understanding task can be formalized as a mapping function:
\begin{equation}
f_\theta: (\mathcal{C}, Q) \rightarrow \mathcal{A}
\end{equation}
where $\mathcal{C}$ represents contextual information (SOP documents, user state, constraints), $Q$ denotes a query or problem scenario, $\mathcal{A}$ is the action space of valid procedural responses, and $\theta$ represents model parameters.

The challenge lies in three interconnected dimensions:

\begin{enumerate}
\item \textbf{Concept Disambiguation:} Given domain terminology $\mathcal{T} = \{t_1, \ldots, t_K\}$, distinguish semantically similar but contextually distinct terms. Formally, for terms $t_i, t_j$ with high similarity, we require:
\begin{equation}
P(t_i|C_1) \neq P(t_j|C_2) \text{ where } C_1 \neq C_2
\end{equation}

\item \textbf{Sequential Dependency Understanding:} For a procedural workflow $W = \langle a_1, a_2, \ldots, a_m \rangle$, verify ordering constraints:
\begin{equation}
\forall i < j: \text{precond}(a_j) \subseteq \text{effect}(a_1) \cup \cdots \cup \text{effect}(a_i)
\end{equation}
where $\text{precond}(a)$ and $\text{effect}(a)$ denote preconditions and effects of action $a$.

\item \textbf{Contextual Constraint Reasoning:} Given a context $c \in \mathcal{C}$ and an action $a \in \mathcal{A}$, evaluate conditional validity:
\begin{equation}
\text{valid}(a|c) = \bigwedge_{i=1}^n \text{constraint}_i(a, c)
\end{equation}
\end{enumerate}

\subsubsection{Limitations of Existing Methods}
Existing methods suffer from three limitations that prevent effective SOP understanding.

First, joint training fails to recognize that procedural understanding requires different types of reasoning capabilities. Distinguishing between samples that require basic concept recognition, sequential reasoning, and contextual reasoning, is essential, yet they treat all samples identically without recognizing that these represent fundamentally different task types.

Second, they lack explicit contrastive signals showing what models should not do. For a correct procedure $W^+ = \langle a_1, \ldots, a_m \rangle$, models never observe systematically constructed incorrect variants $W^-$ that differ minimally from correct workflows yet violate essential procedural constraints, preventing them from learning to distinguish subtle procedural violations across different reasoning task types.

Third, existing evaluation relies on fixed metrics that apply uniform criteria across domains. Such metrics cannot capture domain-specific requirements, e.g., temporal constraint satisfaction in DMV domains, policy compliance verification in banking, or patient safety validation in healthcare, which makes adaptive rubrics  necessary that depend on domain characteristics rather than fixed weighted combinations of surface-level similarity metrics.

\vspace{-2mm}
\subsection{Progressive Task Mixture Framework}
\subsubsection{Three-Stage Architecture}
Our framework structures training through three progressive stages with cumulative data. At stage $k$, the model trains on:
\begin{equation}
\mathcal{D}^{(k)}_{\text{train}} = \bigcup_{i=1}^{k} \mathcal{D}_i
\end{equation}
where $\mathcal{D}_1$ contains concept disambiguation pairs, $\mathcal{D}_2$ adds action sequence pairs, and $\mathcal{D}_3$ incorporates graph reasoning pairs.

\textbf{Difference from Curriculum Learning}: Unlike curriculum learning ordering that samples by difficulty within single task types, our progressive mixture introduces fundamentally different task types at each stage with cumulative data where all previous types remain in training.

\subsubsection{Stage 1: Concept Disambiguation Task}
SOP understanding requires distinguishing domain-specific terminology where lexical similarity does not imply semantic equivalence. For example, in banking SOPs, ``refund'' and ``reimbursement'' may trigger entirely different workflows.

For each concept term $t \in \mathcal{T}$, we generate positive-negative question-answer pairs. Given a query $q$ containing term $t$ and context $c \in \mathcal{C}$, we construct
\begin{equation}
\mathcal{P}_1 = \left\{(q, c, a^+, \{a_j^-\}_{j=1}^m)\right\},
\end{equation}
where $a^+ = f_{\text{correct}}(q, c, t)$ is the correct answer using term $t$ in context $c$, and negative answers $a_j^-$ are generated by substituting semantically similar but contextually inappropriate terms.

\textit{Learning Objective.}
We optimize a contrastive loss
% \begin{equation}
% \mathcal{L}_1(\theta) = -\mathbb{E}_{(q,c,a^+,\{a_j^-\}) \sim \mathcal{P}_1} \left[ \log \frac{\exp(s_\theta(q, c, a^+)/\tau)}{\exp(s_\theta(q, c, a^+)/\tau) + \sum_{j=1}^m \exp(s_\theta(q, c, a_j^-)/\tau)} \right]
% \end{equation}
\begin{equation}
\mathcal{L}_1(\theta) = -\mathbb{E}_{(q,c,a^+,\{a_j^-\}) \sim \mathcal{P}_1}\left[\log\frac{e^{s^+/\tau}}{e^{s^+/\tau}+\sum_{j=1}^m e^{s_j^-/\tau}}\right],
\label{eq:loss}
\end{equation}
where $s^+ = s_\theta(q, c, a^+)$, $s^- = s_\theta(q, c, a^-)$, $s_\theta(q, c, a)$ represents the model's scoring function for answer $a$ given query $q$ and context $c$, and $\tau$ is a temperature parameter.

\subsubsection{Stage 2: Action Sequence Understanding Task}
Procedural correctness requires understanding sequential dependencies and ordering constraints that cannot be captured by concept-level understanding alone. This stage introduces a new task type focused on sequential reasoning.

For a correct workflow $W^+ = \langle a_1, \ldots, a_m \rangle$, we systematically generate negative samples through error injection operators:
\begin{align}
\mathcal{O}_{\text{order}}(W^+) &= \{W' | \exists i < j: W' = \langle \ldots, a_j, \ldots, a_i, \ldots \rangle\}, \\
\mathcal{O}_{\text{omit}}(W^+) &= \{W' | W' = \langle a_1, \ldots, a_{i-1}, a_{i+1}, \ldots, a_m \rangle, i \in [m]\}, \\
\mathcal{O}_{\text{insert}}(W^+) &= \{W' | W' = \langle a_1, \ldots, a_i, a_{\text{spurious}}, a_{i+1}, \ldots \rangle\}.
\end{align}
The complete negative set is
\begin{equation}
\mathcal{W}^- = \mathcal{O}_{\text{order}}(W^+) \cup \mathcal{O}_{\text{omit}}(W^+) \cup \mathcal{O}_{\text{insert}}(W^+).
\label{eq:neg-w}
\end{equation}
The loss is similar to Eq. \ref{eq:loss}, while $s^+ = S_\theta(c, W^+)$ and $s^- = S_\theta(c, W^-)$, $W^- \in \mathcal{W}^-$, compute a sequence-level score for workflow $W$ given context $c$.

% \paragraph{Sequence-Level Contrastive Loss.}
% We extend contrastive learning to sequential outputs: 
% \han{Since we've introduced the contrastive loss before, we do not need to introduce it again. All the three tasks use contrastive loss so we can introduce this loss upfront and then in each subsection we only need to introduce how we construct the negative pairs.}
% \begin{equation}
% \mathcal{L}_2(\theta) = -\mathbb{E}_{(c,W^+,\mathcal{W}^-)} \left[ \log \frac{\exp(S_\theta(c, W^+)/\tau)}{\exp(S_\theta(c, W^+)/\tau) + \sum_{W^- \in \mathcal{W}^-} \exp(S_\theta(c, W^-)/\tau)} \right]
% \end{equation}
% where $S_\theta(c, W)$ computes a sequence-level score for workflow $W$ given context $c$.

\begin{figure*}
    \centering
    \includegraphics[width=0.85\linewidth]{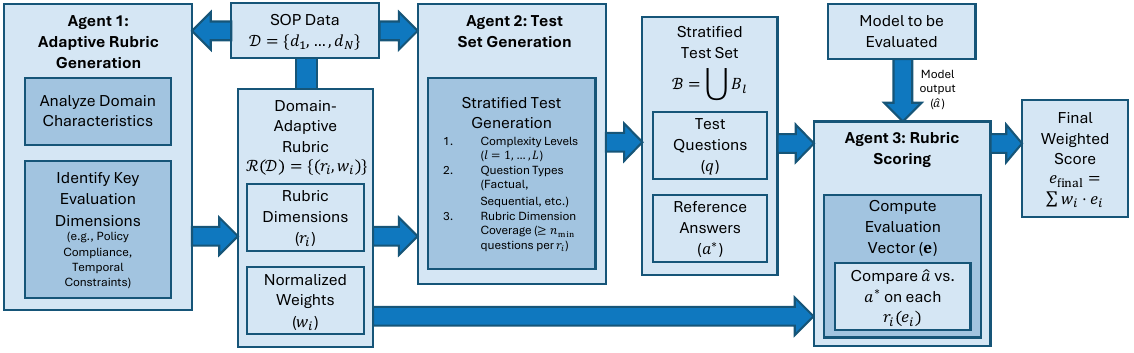}
    \caption{Automatic Multi-Agent Evaluation System. Agent 1 analyzes SOP corpus to generate domain-adaptive rubrics $\mathcal{R}(\mathcal{D}) = \{(r_i, w_i)\}$ with dimensions and weights. Agent 2 creates stratified test set $\mathcal{B}$ across complexity levels, question types, and rubric coverage. Agent 3 computes multi-dimensional scores $\mathbf{e} = [e_1, \ldots, e_{|\mathcal{D}|}]^\top$ and aggregates as $e_{\text{final}} = \sum_i w_i \cdot e_i$, adapting to domain requirements.}
    \label{fig:evaluation_system}
\end{figure*}

\subsubsection{Stage 3: Scenario-Aware Graph Reasoning Task}
Real-world SOP requires reasoning over complex dependencies where actions may have conditional applicability based on context states, temporal constraints, and cross-procedural dependencies. This stage introduces a graph-structured reasoning task type.

We represent each SOP scenario as a Directed Action Graph (DAG):
\begin{equation}
\mathcal{G} = (\mathcal{V}, \mathcal{E}, \phi),
\end{equation}
where $\mathcal{V} = \{v_1, \ldots, v_n\}$ are action nodes, $\mathcal{E} \subseteq \mathcal{V} \times \mathcal{V}$ are directed dependency edges indicating precondition relationships, and $\phi: \mathcal{V} \rightarrow 2^\Phi$ maps each action node to its set of required preconditions from the precondition space $\Phi$. An edge $(v_i, v_j) \in \mathcal{E}$ indicates that action $v_i$ must be completed before $v_j$ can be executed.

Similar to Eq. \ref{eq:neg-w}, We generate negative graphs $\mathcal{G}_{\text{cycle}}^-$, $\mathcal{G}_{\text{precond}}^-$, and $\mathcal{G}_{\text{edge}}^-$ through constraint violations, where $\mathcal{G}_{\text{cycle}}^-$ introduces cyclic dependencies violating execution order, $\mathcal{G}_{\text{precond}}^-$ removes required preconditions for critical actions, and $\mathcal{G}_{\text{edge}}^-$ adds or removes dependency edges creating invalid SOPs. The loss is also similar to Eq. \ref{eq:loss} while $s^+ = R_\theta(c, \mathcal{G}^+)$ and $s^- = R_\theta(c, \mathcal{G}^-)$ combine textual context and graph representations. $R_\theta(c, \mathcal{G})$ represents the model's scoring function for graph $\mathcal{G}$ given context $c$, implemented through structured prompting that presents the graph's nodes, edges, and constraints to the language model.

% \paragraph{Scenario-Aware Contrastive Loss.}
% The final stage objective combines textual context and graph representations:
% \begin{equation}
% \mathcal{L}_3(\theta) = -\mathbb{E}_{(c,\mathcal{G}^+,\{\mathcal{G}_j^-\})} \left[ \log \frac{\exp(R_\theta(c, \mathcal{G}^+)/\tau)}{\exp(R_\theta(c, \mathcal{G}^+)/\tau) + \sum_j \exp(R_\theta(c, \mathcal{G}_j^-)/\tau)} \right]
% \end{equation}
% where $R_\theta(c, \mathcal{G})$ represents the model's scoring function for graph $\mathcal{G}$ given context $c$, implemented through structured prompting that presents the graph's nodes, edges, and constraints to the language model.

\textit{Final Training Objective.}
The complete training loss optimizes
\begin{equation}
\theta^* = \arg\min_\theta \left[ \alpha_1 \mathcal{L}_1(\theta; \mathcal{D}^{(1)}_{\text{train}}) + \alpha_2 \mathcal{L}_2(\theta; \mathcal{D}^{(2)}_{\text{train}}) + \alpha_3 \mathcal{L}_3(\theta; \mathcal{D}^{(3)}_{\text{train}}) \right],
\label{eq:final}
\end{equation}
where $\alpha_1, \alpha_2, \alpha_3 > 0$ are loss weights satisfying $\alpha_1 + \alpha_2 + \alpha_3 = 1$.

\subsection{Automatic Multi-Agent Evaluation System}

% Manual evaluation faces three limitations that necessitate an automatic method. First, human-designed rubrics cannot adapt to domain-specific requirements. Evaluation criteria for banking SOPs (policy compliance, temporal constraints) differ fundamentally from healthcare SOPs (patient safety, privacy compliance), yet fixed rubrics apply uniform standards across domains. Second, the annotation cost is prohibitive: for $M$ models, $N$ test cases, and $K$ evaluation dimensions, manual evaluation requires $O(MNK)$ human judgments, making comprehensive evaluation infeasible for large-scale systems. Third, consistency remains problematic as high variance exists in human judgments, undermining evaluation reliability.
After our progressive task mixture framework for training, 
we now solve the evaluation challenge: how to evaluate SOP understanding 
across diverse domains with varying requirements. Traditional evaluation faces a fundamental limitation: fixed metrics 
apply uniform criteria across all domains, failing to capture that banking 
SOPs require regulatory compliance assessment while healthcare SOPs need 
patient safety verification. We propose a three-agent automatic evaluation system that generates 
adaptive rubrics, stratified test sets, and rubric 
scoring. This enables scalable evaluation 
that automatically adapts to different domains.

\subsubsection{Agent 1: Adaptive Rubric Generation}
The first agent analyzes the SOP corpus $\mathcal{D} = \{d_1, \ldots, d_N\}$ to generate a domain-adaptive rubric $\mathcal{R}(\mathcal{D}) = \{(r_i, w_i)\}_{i=1}^{N}$, where each tuple consists of a rubric dimension $r_i$ and its normalized weight $w_i \in [0, 1]$ with $\sum_i w_i = 1$.

The prompt instructs the agent to identify rubrics (e.g., policy compliance, temporal constraint validation), determining their relative importance through domain characteristic analysis, and specifying concrete scoring criteria for each rubric. This generation ensures that rubrics adapt to domain requirements. For instance, banking SOPs emphasize regulatory compliance and audit trails, while healthcare SOPs prioritize patient safety protocols and privacy protection.

\subsubsection{Agent 2: Test Set Generation}
The second agent generates a stratified test set $\mathcal{B} = \bigcup_{l=1}^L \mathcal{B}_l$ from the SOP corpus $\mathcal{D}$ and rubric $\mathcal{R}(\mathcal{D})$, where each subset $\mathcal{B}_l$ contains test questions at complexity level $l$. We stratify test generation across three dimensions: (1) complexity levels ranging from basic factual recall to multi-step reasoning, (2) question types including factual, sequential, conditional, and multi-hop questions, and (3) rubric dimension coverage ensuring that each dimension $r_i \in \mathcal{R}$ is tested by at least $n_{\min}$ questions.

\subsubsection{Agent 3: Rubric Scoring}
The third agent evaluates model outputs against reference answers using the adaptive rubric. For each test question $(q, a^*) \in \mathcal{B}$ and model output $\hat{a}$, the agent computes an evaluation vector $\mathbf{e} = [e_1, \ldots, e_{N}]^\top$ where each component $e_i$ measures performance on rubric dimension $r_i$ by comparing $\hat{a}$ against $a^*$. 

The reference answers $a^*$ are generated using a separate LLM that is independent from all agents in our framework. We acknowledge that without human annotation, this method is necessary for automatic evaluation at scale. The generated reference answers serve as pseudo ground truth for our evaluation system. We emphasize that the primary validation of our framework comes from experimental results on publicly available benchmarks (\S\ref{sec:main-res}), where we compare against SOTA baselines using standardized test sets. Our automatic evaluation system should be viewed as a functional component that enables scalable, domain-adaptive evaluation, complementing rather than replacing benchmark-based validation.

The final weighted score aggregates across all dimensions:
\begin{equation}
e_{\text{final}} = \sum_{i=1}^{|\mathcal{D}|} w_i \cdot e_i.
\end{equation}

\begin{table*}[!t]
\renewcommand{\arraystretch}{0.85}
\centering
\caption{Performance comparison on SOPBench across seven domains. Pass Rate (\%) reported.}
\label{tab:main_results}
\resizebox{0.85\textwidth}{!}{
\begin{tabular}{lccccccc|c}
\toprule
\textbf{Model} & \textbf{Bank} & \textbf{DMV} & \textbf{Healthcare} & \textbf{Market} & \textbf{Univ} & \textbf{Library} & \textbf{Hotel} & \textbf{Overall} \\
\midrule
\multicolumn{9}{c}{\textit{Proprietary Non-reasoning Models}} \\
\midrule
GPT-4.1 (FC) & 71.89 & 78.26 & 80.00 & 81.82 & 52.27 & 61.54 & 42.86 & 67.22 \\
GPT-4o (FC) & 64.71 & 80.87 & 73.85 & 63.64 & 68.18 & 65.38 & 39.80 & 62.13 \\
Claude-3-7-Sonnet (FC) & 69.28 & 70.43 & 72.31 & 58.29 & 68.18 & 37.18 & 23.98 & 54.26 \\
GPT-4.1-mini (FC) & 62.75 & 73.91 & 67.69 & 58.82 & 38.64 & 25.64 & 7.65 & 47.07 \\
Claude-3-5-Sonnet (FC) & 71.90 & 50.43 & 39.23 & 43.32 & 52.27 & 33.33 & 15.82 & 41.42 \\
GPT-4o-mini (FC) & 34.64 & 70.43 & 26.15 & 45.99 & 40.91 & 46.15 & 41.33 & 42.64 \\
Gemini-2.0-Flash (FC) & 56.86 & 54.78 & 23.08 & 40.11 & 34.09 & 26.92 & 7.65 & 33.33 \\
Gemini-1.5-Pro (FC) & 54.25 & 60.00 & 18.46 & 34.22 & 63.64 & 26.92 & 12.37 & 34.18 \\
\midrule
\multicolumn{9}{c}{\textit{Reasoning Models}} \\
\midrule
o4-mini-high (FC) & 76.47 & 81.74 & 93.08 & 90.37 & 95.45 & 43.59 & 56.12 & 76.08 \\
Gemini-2.0-Flash-Thinking (ReAct) & 77.12 & 73.91 & 83.08 & 53.48 & 93.18 & 55.13 & 62.24 & 67.66 \\
Deepseek-R1 (ReAct) & 55.56 & 79.13 & 55.38 & 71.66 & 77.27 & 57.69 & 51.02 & 62.13 \\
Claude-3.7-Sonnet-Thinking (FC) & 71.90 & 72.17 & 73.85 & 50.80 & 70.45 & 34.62 & 23.47 & 53.27 \\
\midrule
\multicolumn{9}{c}{\textit{Open-source Models}} \\
\midrule
Llama-3.1-70B-Instruct (ReAct) & 43.79 & 66.96 & 50.96 & 40.44 & 45.45 & 42.86 & 14.29 & 41.20 \\
Qwen-2.5-72B-Instruct (ReAct) & 32.68 & 61.74 & 28.46 & 41.71 & 38.64 & 38.46 & 14.29 & 34.44 \\
Qwen-2.5-32B-Instruct (ReAct) & 41.83 & 53.04 & 42.31 & 46.52 & 56.82 & 37.18 & 18.88 & 39.65 \\
Qwen-2.5-14B-Instruct (ReAct) & 32.03 & 53.91 & 29.23 & 39.04 & 27.27 & 30.77 & 15.31 & 31.89 \\
Llama-3.1-8B-Instruct (ReAct) & 13.73 & 20.00 & 20.00 & 19.25 & 25.00 & 32.05 & 0.51 & 15.84 \\
Qwen-2.5-7B-Instruct (ReAct) & 5.88 & 21.74 & 17.69 & 13.37 & 2.27 & 21.79 & 1.02 & 11.30 \\
\midrule
\multicolumn{9}{c}{\textit{FM SO.P (Ours)}} \\
\midrule
\textbf{FM SO.P (Qwen-2.5-7B-Instruct)} & \textbf{29.85} & \textbf{38.14} & \textbf{41.13} & \textbf{31.98} & \textbf{45.24} & \textbf{27.27} & \textbf{26.67} & \textbf{34.33} \\
\textbf{FM SO.P (Qwen-2.5-14B-Instruct)} & \textbf{42.54} & \textbf{63.92} & \textbf{41.13} & \textbf{26.16} & \textbf{35.71} & \textbf{33.33} & \textbf{34.87} & \textbf{39.67} \\
\textbf{FM SO.P (Qwen-2.5-32B-Instruct)} & \textbf{58.21} & \textbf{59.79} & \textbf{41.94} & \textbf{46.51} & \textbf{52.38} & \textbf{45.45} & \textbf{33.85} & \textbf{48.30} \\
\bottomrule
\end{tabular}
}
\end{table*}

\section{Experiments}

\subsection{Experimental Setup}

\subsubsection{Datasets and Domains}
We evaluate our framework on SOPBench~\cite{li2025sopbench}, a comprehensive benchmark for evaluating language agents on SOPs. SOPBench covers seven diverse domains: Banking (Bank), Department of Motor Vehicles (DMV), Healthcare (Health), Marketplace (Market), University Administration (Univ), Library Services (Lib), and Hotel Management (Hotel), with verified test cases spanning multiple complexity levels and constraint types.

For each domain, we construct training data $\mathcal{D}_{\text{train}} = \mathcal{D}_1 \cup \mathcal{D}_2 \cup \mathcal{D}_3$ following the three-stage task mixture design: concept disambiguation task pairs ($\mathcal{D}_1$), action sequence task pairs with procedural errors ($\mathcal{D}_2$), and graph-structured scenario reasoning task pairs ($\mathcal{D}_3$).

\subsubsection{Implementation Details}
We implement FM SO.P based on Qwen-2.5 series models: Qwen-2.5-7B, Qwen-2.5-14B, and Qwen-2.5-32B. Training is conducted using the unified objective in Eq. \ref{eq:final} with loss weights $\alpha_1 = 1/3$, $\alpha_2 = 1/3$, $\alpha_3 = 1/3$. We use AdamW optimizer with learning rate $1 \times 10^{-4}$, batch size 256, and train for 12 epochs. Temperature parameter $\tau = 1$ for all contrastive losses. All experiments are conducted on $8\times$ NVIDIA H100-80GB GPUs with mixed-precision training (bfloat16).

\subsubsection{Baseline Methods}
We compare FM SO.P with baseline models from three categories, following the SOPBench evaluation protocol. The first category includes \textit{proprietary non-reasoning models}: GPT-4.1 (FC), GPT-4o (FC), Claude-3-7-Sonnet (FC), GPT-4.1-mini (FC), Claude-3-5-Sonnet (FC), GPT-4o-mini (FC), Gemini-2.0-Flash (FC), and Gemini-1.5-Pro (FC), where FC indicates function-calling capability. The second category consists of \textit{reasoning models} that use explicit reasoning mechanisms such as ReAct or extended thinking processes, including o4-mini-high (FC), Gemini-2.0-Flash-Thinking (ReAct), Deepseek-R1 (ReAct), and Claude-3.7-Sonnet-Thinking (FC). The third category includes \textit{open-source models} representing state-of-the-art alternatives with various parameter scales: Llama-3.1-70B-Instruct (ReAct), Qwen2.5-7B-Instruct (ReAct), Qwen-2.5-32B-Instruct (ReAct), Qwen-2.5-14B-Instruct (ReAct), Llama-3.1-8B-Instruct (ReAct), and Qwen-2.5-7B-Instruct (ReAct). For fair comparison, we apply our FM SO.P training framework to the same base architectures used in open-source baselines (Qwen-2.5-7B, Qwen-2.5-14B, and Qwen-2.5-32B).

\subsubsection{Evaluation Metrics}
We evaluate model performance using two main metrics. \textit{Pass rate} measures the percentage of test cases where the model output correctly solves the SOP task according to ground truth action sequences, providing a standardized measure for comparing different models. This metric captures whether models can generate procedurally correct, complete, and properly ordered action sequences that satisfy all task requirements. We also compute \textit{Borda count} \cite{de1995elections,siththaranjandistributional,ghosh2025onebench} based on group performance comparisons. Across all test cases, we determine which method produces better outputs, and the Borda count aggregates these win rates to produce a group ranking that accounts for relative performance rather than absolute scores alone. This metric is particularly valuable when comparing methods with different inference strategies or when performance distributions overlap significantly.

\begin{figure*}[t]
    \centering
    \includegraphics[width=0.85\linewidth]{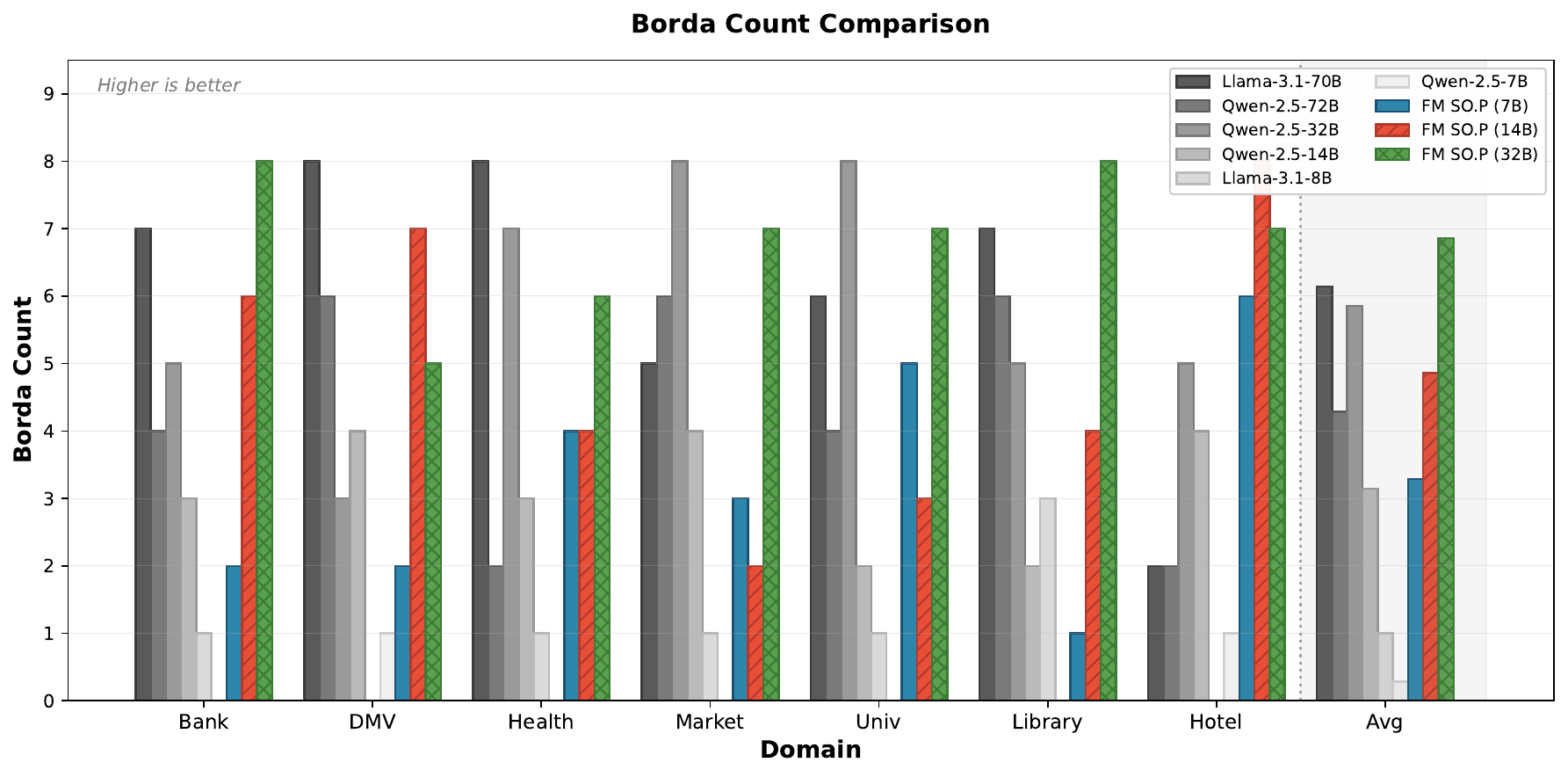}
    \vspace{-1em}
    \caption{\textbf{Borda counts for models across domains.} Higher score is better. FM SO.P variants trained with progressive task mixture achieve superior quality across all model sizes.}
    \label{fig:condorcet_all}
\end{figure*}

\subsection{Main Results}
\label{sec:main-res}

\subsubsection{Overall Performance Comparison}

Table~\ref{tab:main_results} shows comparison between FM SO.P and baseline models across all seven domains. FM SO.P (Qwen-2.5-7B) achieves 34.33\% overall pass rate, representing a +23.03\% improvement over the base Qwen-2.5-7B-Instruct model (11.30\%). The 14B variant achieves 39.67\% (+7.78\% over its 31.89\% base), and most notably, the 32B variant reaches 48.30\% (+8.65\% over its 39.65\% base), establishing new state-of-the-art performance among open-source models. Our results demonstrate the effectiveness of progressive task mixture for SOP understanding, with consistent improvements scaling across model sizes.

% \paragraph{Analysis of Results.}
% The results provide compelling evidence for the effectiveness of our progressive task mixture framework across multiple dimensions. While the 7B model shows the largest relative gain due to starting from an extremely weak base (11.30\%), the 32B model achieves the strongest absolute performance (48.30\%), surpassing the much larger Qwen-2.5-72B-Instruct baseline (34.44\%) by +13.86\% despite using 2.25$\times$ fewer parameters. This demonstrates that structured procedural training with task mixture can be more impactful than scale alone.

Comparing the parameter efficiency, FM SO.P (Qwen-2.5-7B) matches Qwen-2.5-72B-Instruct performance (34.33\% vs 34.44\%) with 10$\times$ fewer parameters. FM SO.P (Qwen-2.5-32B) largely outperforms Qwen-2.5-72B-Instruct and Llama-3.1-70B-Instruct (48.30\% vs 34.44\%/41.20\%, +13.86\%/7.10\%) while using 2.25/2.2$\times$ fewer parameters, and achieves competitive performance with reasoning models like Claude-3.7-Sonnet-Thinking (53.27\%) without inference-time reasoning mechanisms. This parameter efficiency validates that task mixture design can compensate for scale disadvantages.

% \vspace{-0.5em}

\subsubsection{Borda Count Analysis}
We show Borda count analysis comparing all models to evaluate relative output quality through group comparisons. Fig.~\ref{fig:condorcet_all} shows domain-wise counts where higher numbers indicate better performance. FM SO.P (32B) raises from 5.86 (base Qwen-2.5-32B) to 6.86 (+1.00), achieving the highest count while Llama-3.1-70B/Qwen-2.5-72B only achieves 6.14/4.29. FM SO.P (7B) raises from 0.29 (base Qwen-2.5-7B) to 3.29 (+3.00), representing a 11.5$\times$ improvement, transforming a model with near-zero preference into one matching Qwen-2.5-14B (3.14) despite 2$\times$ fewer parameters. FM SO.P (14B) raises from 3.14 (base) to 4.86 (+1.72), surpassing Qwen-2.5-72B (4.29) with 5$\times$ fewer parameters. The Borda counts validates that task mixture not only improves task pass rates but also improves output quality. These results also suggest that task mixture provides greater relative quality benefits when base models are weaker.

\subsection{Ablation Studies}

\subsubsection{Impact of Task Mixture Stages.}
We separately add training stages to validate the effectiveness of our progressive task mixture (Table~\ref{tab:stage_ablation}). The overall +23.03\% improvement (11.30\% $\to$ 34.33\%) surpasses any single stage's contribution, demonstrating that the three stages systematically build SOP understanding and have positive interrelation. Stage 1 establishes terminology foundation (average +5.81\%), with significant gains in Healthcare and Bank (+22.63\%, +11.28\%) validating that concept disambiguation effectively handles terminology. Stage 2 builds sequential reasoning through procedural error injection (+7.39\%), with Healthcare jumping to 51.61\% (+11.29\%) showing that action sequence successfully captures ordering constraints. Stage 3 brings the largest improvement (+9.83\%), with University's large improvement (+33.34\%) proving that graph reasoning effectively handles conditional logic. 

Different domain trajectories reveal how progressive mixture adapts to domains. Sequential domains (Healthcare, Bank) obtain major increases from Stage 1-2 (Healthcare: +33.92\% cumulative from S1+S2), while graph domains (University, Market, Library) mainly increase in Stage 3 (University: +33.34\% from S3 alone). Balanced domains (DMV, Hotel) benefit uniformly from all stages.

\begin{table*}[t]
\renewcommand{\arraystretch}{0.85}
\centering
\caption{\textbf{Progressive task mixture stage ablation.} Pass Rate (\%) across domains. Each row adds one training stage (task type) to the previous stage, with cumulative data.}
\label{tab:stage_ablation}
\resizebox{0.85\textwidth}{!}{
\begin{tabular}{lcccccccc}
\toprule
\textbf{Model Configuration} & \textbf{Bank} & \textbf{DMV} & \textbf{Healthcare} & \textbf{Market} & \textbf{Univ} & \textbf{Library} & \textbf{Hotel} & \textbf{Overall} \\
\midrule
% \multicolumn{9}{c}{\textit{Qwen3-8B Base Model}} \\
% \midrule
% Qwen3-8B (ReAct) & 30.60 & 32.99 & 15.32 & 10.47 & 26.10 & 22.73 & 13.85 & 21.73 \\
% + Stage 1 (Concept Disambiguation) & 31.34 & 37.11 & 30.65 & 20.35 & 21.43 & 19.70 & 14.87 & 25.06 \\
% \midrule
% $\Delta$ from base & +0.74 & +4.12 & +15.33 & +9.88 & -4.67 & -3.03 & +1.02 & +3.33 \\
% \midrule
% \multicolumn{9}{c}{\textit{Qwen2.5-7B Base Model}} \\
% \midrule
Qwen2.5-7B-Instruct (ReAct) & 5.88 & 21.74 & 17.69 & 13.37 & 2.27 & 21.79 & 1.02 & 11.30 \\
+ Stage 1 (Concept Disambiguation) & 17.16 & 24.74 & 40.32 & 13.95 & 2.38 & 21.21 & 0.00 & 17.11 \\
+ Stage 2 (Action Sequence) & 33.58 & 34.02 & 51.61 & 14.53 & 11.90 & 19.70 & 6.15 & 24.50 \\
+ Stage 3 (Graph Reasoning) & 29.85 & 38.14 & 41.13 & 31.98 & 45.24 & 27.27 & 26.67 & 34.33 \\
\midrule
$\Delta$ Stage 1 & +11.28 & +3.00 & +22.63 & +0.58 & +0.11 & -0.58 & -1.02 & +5.81 \\
$\Delta$ Stage 2 & +16.42 & +9.28 & +11.29 & +0.58 & +9.52 & -1.51 & +6.15 & +7.39 \\
$\Delta$ Stage 3 & -3.73 & +4.12 & -10.48 & +17.45 & +33.34 & +7.57 & +20.52 & +9.83 \\
$\Delta$ Total & +23.97 & +16.40 & +23.44 & +18.61 & +42.97 & +5.48 & +25.65 & +23.03 \\
\bottomrule
\end{tabular}
}
\end{table*}

\begin{figure}
    \centering
    \includegraphics[width=\linewidth]{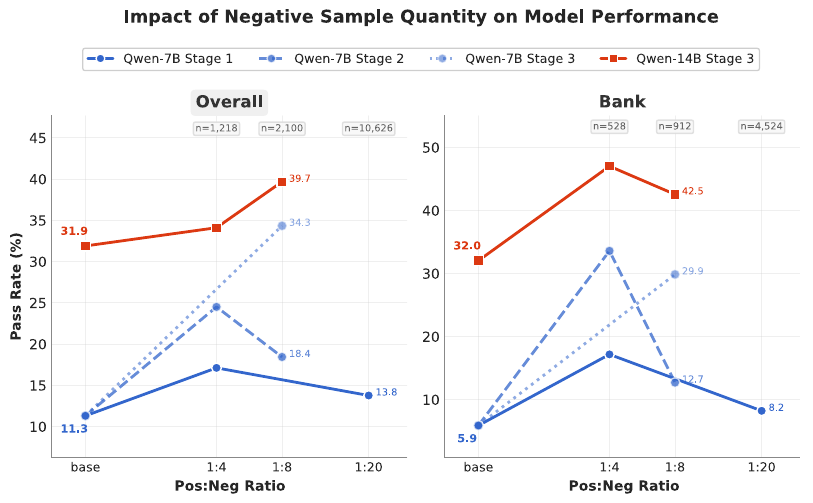}
    \vspace{-3em}
    \caption{\textbf{Negative sample ratio ablation across task mixture stages.} Pass Rate (\%) with varying positive:negative ratios.}
    \label{fig:negative_ablation}
\end{figure}

\subsubsection{Impact of Negative Sample Ratio.}
Fig.~\ref{fig:negative_ablation} shows performance across varying positive-to-negative ratios (1:4, 1:8, 1:20) for our three training stages. For the 7B model, Stage 1 (concept disambiguation) performs best at 1:4 ratio, Stage 2 (action sequence understanding) maintains best at 1:4, while Stage 3 (graph reasoning) achieves highest performance at 1:8. The 14B model at Stage 3 shows similar pattern with 1:8 as the best ratio. This reflects increasing representation complexity across training stages. We also have domain-specific analysis to show how different task types benefit from varying negative ratios. For example, banking domain shows Stage 1 highest at 1:4 (17.2\%), with significant drops at 1:20 (8.2\%). For detailed domain results, see Appendix.

Generally, for task mixture design, best contrastive learning ratios are task-dependent, not universal. The common practice of using fixed negative ratios across all training stages is not the best strategy. Our results demonstrate that task mixture should use adaptive negative sampling, adjusting ratios based on the representation complexity of each stage's learning objective. The best ratios also provide insight into why FM SO.P succeeds. By using 1:4 for Stages 1-2 and 1:8 for Stage 3, we optimize each stage's contrastive learning effectiveness.

\begin{figure*}
    \centering
    \includegraphics[width=\linewidth]{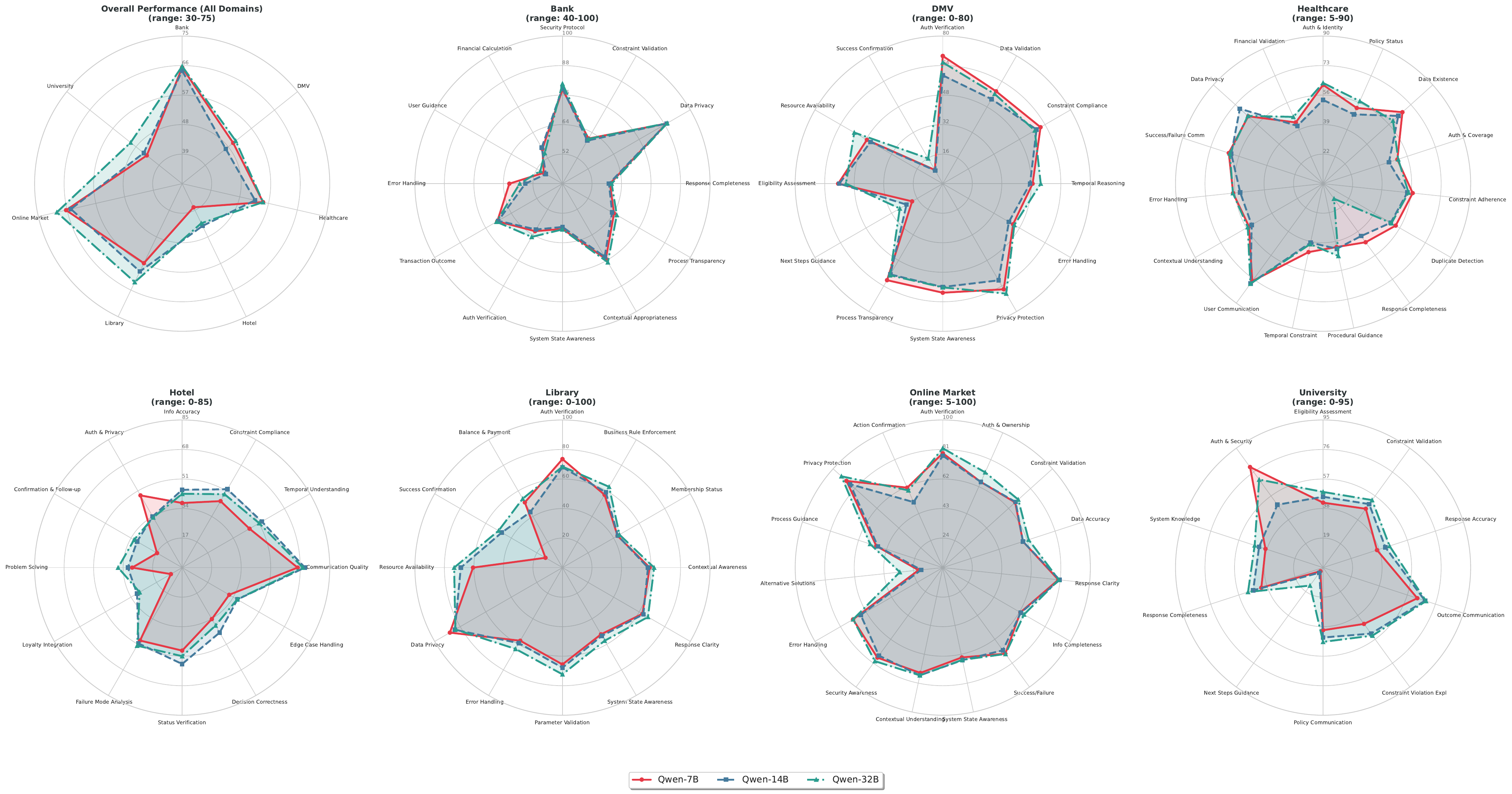}
    \caption{Domain-Adaptive Rubric Analysis. Radar plots showing FM SO.P performance across generated rubrics for seven domains. Each domain has unique rubric dimensions (e.g., Banking: security protocol and financial accuracy; DMV: temporal reasoning and eligibility). Blue (7B), orange (14B), and green (32B) show scaling effects across model sizes.}
    \label{fig:radar}
\end{figure*}

\subsection{Domain-Adaptive Evaluation Analysis}

We analyze performance across domain-adaptive rubrics generated by our multi-agent evaluation system. Fig.~\ref{fig:radar} shows radar plots of model performance across rubrics for all seven domains, with each domain rubric is generated by Agent 1. For example, Banking SOPs has 12 rubrics including security protocol adherence (78.3--80.6\% across model sizes) and financial calculation accuracy (54.4--56.9\%). In contrast, DMV SOPs focus on temporal reasoning (47.3--53.1\%) and eligibility assessment (52.5--56.7\%), reflecting the time-sensitive nature of government services. This rubric generation validates varying SOP requirements across domains rather than uniform metrics.

Certain rubrics show large improvements with increased model size, e.g., authentication verification (DMV: 58.7\% → 69.2\%, +10.5\%), resource availability assessment (Library: 60.6\% → 73.4\%, +12.8\%), alternative solutions (Online Market: 19.0\% → 32.8\%, +13.8\%), and loyalty program integration (Hotel: 7.5\% → 30.0\%, +22.5\%). These dimensions involve multi-step reasoning over resource states and eligibility conditions that become more robust with scale. From these increase, Our 7B models are good at core SOP capabilities (privacy, authentication, basic constraint reasoning). Our 14B models provide balanced performance across most rubrics, offering a sweet spot for general-purpose SOP understanding. Our 32B models are good for complex scenarios (generating alternative solutions, managing tiered service programs, coordinating multi-step interactions).

\subsection{Case Study: Generated Domain SOP}

To demonstrate the practical capability of our framework, we show a
representative generated Healthcare SOP. This illustrates how FM SO.P captures domain-specific procedural
knowledge, conditional logic, and constraint requirements. For other cases, see Appendix.

\textit{Scenario.}
A patient seeks to appeal a denied insurance claim through the healthcare provider's system.

% \subsubsection{SOP}

\paragraph{Step 1: User Authentication}
\textbf{Action:} Verify user identity

\begin{lstlisting}[language=Python, basicstyle=\ttfamily\small]
CALL: login_user(username, identification)
\end{lstlisting}

\textbf{Decision Point 1:} Is user authenticated?
\begin{itemize}
    \item \textbf{IF NO} $\rightarrow$ \textbf{TERMINATE}: Return error ``Authentication Failed''
    \item \textbf{IF YES} $\rightarrow$ \textbf{PROCEED} to Step 2
\end{itemize}

\paragraph{Step 2: Policy Verification}
\textbf{Action:} Validate policy status

\begin{lstlisting}[language=Python, basicstyle=\ttfamily\small]
CALL: get_policy_details(username)
AND: policy_active(username)
\end{lstlisting}

\textbf{Decision Point 2:} Is policy currently active?
\begin{itemize}
    \item \textbf{IF NO} $\rightarrow$ \textbf{TERMINATE}: Return error ``Policy Inactive''
    \item \textbf{IF YES} $\rightarrow$ \textbf{PROCEED} to Step 3
\end{itemize}

\paragraph{Steps 3--4: Intermediate Verification}
The complete workflow includes additional verification steps for claim existence and appeal timeframe validation. These intermediate decision points follow the same pattern of conditional branching with early termination on failure. See Appendix for complete details.

\paragraph{Step 5: Appeal Processing}
\textbf{Action:} Create appeal record and initiate review

\begin{lstlisting}[language=Python, basicstyle=\ttfamily\small]
EXECUTE:
  - Generate appeal_id
  - Set status = "PENDING_REVIEW"
  - Assign to review queue
  - Send confirmation to user
\end{lstlisting}

\textbf{Outcome:} \textsc{Success} - Appeal submitted for review

\section{Related Work}

\textit{SOP Understanding.}
Standard Operating Procedures encode critical organizational knowledge, yet existing language models struggle with procedural reasoning \cite{brahman2023plasma, ruisprocedural, shen2024taskbench}. Recent benchmarks reveal significant gaps: SOP-Bench \cite{nandi2025sop} shows 27-48\% success rates across 1,800+ tasks, while SOPBench \cite{li2025sopbench} demonstrates 30-50\% pass rates with 167 tools across 7 domains. However, these benchmarks rely on fixed evaluation metrics that apply uniform criteria, failing to capture domain-specific requirements (e.g., temporal constraints for DMV vs. regulatory compliance for banking). Moreover, no prior work systematically solves how to train models for SOP through specific task design.

\textit{Progressive Training and Task Mixture.}
Curriculum learning \cite{bengio2009curriculum, soviany2022curriculum, wang2021survey} orders training samples by difficulty using predefined metrics. Recent methods include GPT-4-scored stratification \cite{pang2024phased} and attention-based ordering \cite{kim2024strategic}. However, these methods order samples within single task types and replace easier data with harder data. FM SO.P differs fundamentally by introducing different task types at each stage (concept $\to$ sequence $\to$ graph) with cumulative data, optimizing across multiple task-specific objectives that build hierarchically rather than ordering by difficulty alone.

\textit{Automatic Evaluation Systems.}
Multi-dimensional evaluation methods like LLM-Rubric \cite{hashemi2024llm}, FollowBench \cite{jiang2024followbench}, and Prometheus \cite{kim2023prometheus} establish LLM-as-judge but rely on fixed rubrics or manual construction. Recent multi-agent systems \cite{wu2024autogen, hong2023metagpt} demonstrate collaboration, yet no prior work applies this to adaptive SOP evaluation. FM SO.P's three-agent system adaptively generates domain rubrics, stratified test sets, and rubric scoring without human annotation, enabling scalable evaluation that adapts to domain requirements.

% \section{Conclusion}

% We propose FM SO.P, a framework for cross-domain SOP understanding through progressive task mixture and automatic evaluation. Our three-stage task mixture trains model through contrastive pair generation across three different task types with increasing complexity, from basic terminology precision (concept disambiguation) to sequential correctness (action sequence understanding) to complex conditional reasoning (graph reasoning). Our multi-agent evaluation system generates adaptive rubrics that capture varying requirements across domains, enabling scalable, context-aware evaluation.

% This method achieves large improvements across seven diverse domains, with FM SO.P (Qwen-2.5-7B) improving from 11.30\% to 34.33\% overall pass rate (+23.03\%), matching Qwen-2.5-72B-Instruct performance with 10× fewer parameters. The 14B variant reaches 39.67\%, and the 32B variant achieves 48.30\%, establishing new state-of-the-art among open-source models.

% Our method establishes that progressive task mixture can effectively teach complex procedural reasoning, with impacts extending beyond SOPs to any domain requiring procedural workflow and contextual constraint. The framework's ability to match large-model performance with significantly fewer parameters makes enterprise-scale SOP automation practical, opening pathways for automatic compliance verification, workflow optimization, and intelligent process assistance.

\section{Conclusion}
We propose FM SO.P, a framework for cross-domain SOP understanding through progressive task mixture and automatic evaluation. Our three-stage task mixture trains models through contrastive pair generation across different task types with increasing complexity: from terminology precision (concept disambiguation) to sequential correctness (action sequence understanding) to conditional reasoning (graph reasoning). Our multi-agent evaluation system generates adaptive rubrics capturing domain-specific requirements, enabling scalable evaluation. Our method achieves large improvements across seven domains, with FM SO.P (7B) improving from 11.30\% to 34.33\% (+23.03\%), matching 72B baseline with 10× fewer parameters. The 14B and 32B variants reach 39.67\% and 48.30\%, establishing new state-of-the-art among open-source models.

Our method demonstrates that progressive task mixture effectively teaches complex procedural reasoning, with potential extending to any domain requiring procedural workflows and contextual constraints. The ability to match large-model performance with significantly fewer parameters makes enterprise-scale SOP automation practical, enabling automatic compliance verification, workflow optimization, and intelligent process assistance.

\section{Acknowledgments}
This paper was conducted during an internship at Amazon and supported by the Perfect Order Experience (PoE) team. We are grateful for the computation resources provided and the collaborative research environment that enabled this research. We would like to thank Vincent Gao for his mentorship and guidance. Special thanks to Yang Liu for his valuable suggestions and insights throughout the project. We also extend our appreciation to Lingdao Sha, Jinfeng Xiao, and all other members of the PoE team for their ongoing feedback, thoughtful discussions, and contributions to improving this work.

\bibliographystyle{ACM-Reference-Format}
\bibliography{reference}

@article{li2025sopbench,
  title={SOPBench: Evaluating Language Agents at Following Standard Operating Procedures and Constraints},
  author={Li, Zekun and Huang, Shinda and Wang, Jiangtian and Zhang, Nathan and Antoniades, Antonis and Hua, Wenyue and Zhu, Kaijie and  Zeng, Sirui and Wang, Chi and Wang, William Yang and Yan, Xifeng},
  journal={arXiv preprint arXiv:2506.08669},
  year={2025}
}

@article{nandi2025sop,
  title={SOP-Bench: Complex Industrial SOPs for Evaluating LLM Agents},
  author={Nandi, Subhrangshu and Datta, Arghya and Vichare, Nikhil and Bhattacharya, Indranil and Raja, Huzefa and Xu, Jing and Ray, Shayan and Carenini, Giuseppe and Srivastava, Abhi and Chan, Aaron and others},
  journal={arXiv preprint arXiv:2506.08119},
  year={2025}
}

@inproceedings{pang2024phased,
  title={Phased Instruction Fine-Tuning for Large Language Models},
  author={Pang, Wei and Zhou, Chuan and Zhou, Xiao-Hua and Wang, Xiaojie},
  booktitle={Findings of the Association for Computational Linguistics ACL 2024},
  pages={5735--5748},
  year={2024}
}

@article{kim2024strategic,
  title={Strategic data ordering: Enhancing large language model performance through curriculum learning},
  author={Kim, Jisu and Lee, Juhwan},
  journal={arXiv preprint arXiv:2405.07490},
  year={2024}
}

@inproceedings{hashemi2024llm,
  title={LLM-Rubric: A Multidimensional, Calibrated Approach to Automated Evaluation of Natural Language Texts},
  author={Hashemi, Helia and Eisner, Jason and Rosset, Corby and Van Durme, Benjamin and Kedzie, Chris},
  booktitle={Proceedings of the 62nd Annual Meeting of the Association for Computational Linguistics (Volume 1: Long Papers)},
  pages={13806--13834},
  year={2024}
}

@inproceedings{jiang2024followbench,
  title={Followbench: A multi-level fine-grained constraints following benchmark for large language models},
  author={Jiang, Yuxin and Wang, Yufei and Zeng, Xingshan and Zhong, Wanjun and Li, Liangyou and Mi, Fei and Shang, Lifeng and Jiang, Xin and Liu, Qun and Wang, Wei},
  booktitle={Proceedings of the 62nd Annual Meeting of the Association for Computational Linguistics (Volume 1: Long Papers)},
  pages={4667--4688},
  year={2024}
}

@inproceedings{brahman2024plasma,
  title={Plasma: Procedural knowledge models for language-based planning and re-planning},
  author={Brahman, Faeze and Bhagavatula, Chandra and Pyatkin, Valentina and Hwang, Jena D and Li, Xiang Lorraine and Arai, Hirona Jacqueline and Sanyal, Soumya and Sakaguchi, Keisuke and Ren, Xiang and Choi, Yejin},
  booktitle={The Twelfth International Conference on Learning Representations},
  year={2024}
}

@article{kulkarni2025agent,
  title={Agent-s: Llm agentic workflow to automate standard operating procedures},
  author={Kulkarni, Mandar},
  journal={arXiv preprint arXiv:2503.15520},
  year={2025}
}

@article{garg2025generating,
  title={Generating Structured Plan Representation of Procedures with LLMs},
  author={Garg, Deepeka and Zeng, Sihan and Ganesh, Sumitra and Ardon, Leo},
  journal={arXiv preprint arXiv:2504.00029},
  year={2025}
}

@article{wang2025sop,
  title={SOP-Maze: Evaluating Large Language Models on Complicated Business Standard Operating Procedures},
  author={Wang, Jiaming and Tang, Zhe and Jin, Yilin and Ding, Peng and Li, Xiaoyu and Cao, Xuezhi},
  journal={arXiv preprint arXiv:2510.08942},
  year={2025}
}

@inproceedings{grohs2023large,
  title={Large language models can accomplish business process management tasks},
  author={Grohs, Michael and Abb, Luka and Elsayed, Nourhan and Rehse, Jana-Rebecca},
  booktitle={International conference on business process management},
  pages={453--465},
  year={2023},
  organization={Springer}
}

@inproceedings{xiao2024flowbench,
  title={FlowBench: Revisiting and Benchmarking Workflow-Guided Planning for LLM-based Agents},
  author={Xiao, Ruixuan and Ma, Wentao and Wang, Ke and Wu, Yuchuan and Zhao, Junbo and Wang, Haobo and Huang, Fei and Li, Yongbin},
  booktitle={Findings of the Association for Computational Linguistics: EMNLP 2024},
  pages={10883--10900},
  year={2024}
}

@article{yao2024tau,
  title={$\backslash tau $-bench: A Benchmark for Tool-Agent-User Interaction in Real-World Domains},
  author={Yao, Shunyu and Shinn, Noah and Razavi, Pedram and Narasimhan, Karthik},
  journal={arXiv preprint arXiv:2406.12045},
  year={2024}
}

@article{brahman2023plasma,
  title={Plasma: Making small language models better procedural knowledge models for (counterfactual) planning},
  author={Brahman, Faeze and Bhagavatula, Chandra and Pyatkin, Valentina and Hwang, Jena D and Li, Xiang Lorraine and Arai, Hirona J and Sanyal, Soumya and Sakaguchi, Keisuke and Ren, Xiang and Choi, Yejin},
  journal={arXiv preprint arXiv:2305.19472},
  year={2023}
}

@article{diao2025guidebench,
  title={GuideBench: Benchmarking Domain-Oriented Guideline Following for LLM Agents},
  author={Diao, Lingxiao and Xu, Xinyue and Sun, Wanxuan and Yang, Cheng and Zhang, Zhuosheng},
  journal={arXiv preprint arXiv:2505.11368},
  year={2025}
}

@article{ouyang2022training,
  title={Training language models to follow instructions with human feedback},
  author={Ouyang, Long and Wu, Jeffrey and Jiang, Xu and Almeida, Diogo and Wainwright, Carroll and Mishkin, Pamela and Zhang, Chong and Agarwal, Sandhini and Slama, Katarina and Ray, Alex and others},
  journal={Advances in neural information processing systems},
  volume={35},
  pages={27730--27744},
  year={2022}
}

@inproceedings{weifinetuned,
  title={Finetuned Language Models are Zero-Shot Learners},
  author={Wei, Jason and Bosma, Maarten and Zhao, Vincent and Guu, Kelvin and Yu, Adams Wei and Lester, Brian and Du, Nan and Dai, Andrew M and Le, Quoc V},
  booktitle={International Conference on Learning Representations}
}

@article{chung2024scaling,
  title={Scaling instruction-finetuned language models},
  author={Chung, Hyung Won and Hou, Le and Longpre, Shayne and Zoph, Barret and Tay, Yi and Fedus, William and Li, Yunxuan and Wang, Xuezhi and Dehghani, Mostafa and Brahma, Siddhartha and others},
  journal={Journal of Machine Learning Research},
  volume={25},
  number={70},
  pages={1--53},
  year={2024}
}

@inproceedings{wang2023self,
  title={Self-instruct: Aligning language models with self-generated instructions},
  author={Wang, Yizhong and Kordi, Yeganeh and Mishra, Swaroop and Liu, Alisa and Smith, Noah A and Khashabi, Daniel and Hajishirzi, Hannaneh},
  booktitle={Proceedings of the 61st annual meeting of the association for computational linguistics (volume 1: long papers)},
  pages={13484--13508},
  year={2023}
}

@inproceedings{ruisprocedural,
  title={Procedural Knowledge in Pretraining Drives Reasoning in Large Language Models},
  author={Ruis, Laura and Mozes, Maximilian and Bae, Juhan and Kamalakara, Siddhartha Rao and Gnaneshwar, Dwaraknath and Locatelli, Acyr and Kirk, Robert and Rockt{\"a}schel, Tim and Grefenstette, Edward and Bartolo, Max},
  booktitle={The Thirteenth International Conference on Learning Representations}
}

@article{du2023shortcut,
  title={Shortcut learning of large language models in natural language understanding},
  author={Du, Mengnan and He, Fengxiang and Zou, Na and Tao, Dacheng and Hu, Xia},
  journal={Communications of the ACM},
  volume={67},
  number={1},
  pages={110--120},
  year={2023},
  publisher={ACM New York, NY, USA}
}

@inproceedings{tang2023large,
  title={Large language models can be lazy learners: Analyze shortcuts in in-context learning},
  author={Tang, Ruixiang and Kong, Dehan and Huang, Longtao and others},
  booktitle={Findings of the Association for Computational Linguistics: ACL 2023},
  pages={4645--4657},
  year={2023}
}

@article{sakai2025revisiting,
  title={Revisiting Compositional Generalization Capability of Large Language Models Considering Instruction Following Ability},
  author={Sakai, Yusuke and Kamigaito, Hidetaka and Watanabe, Taro},
  journal={arXiv preprint arXiv:2506.15629},
  year={2025}
}

@inproceedings{wanggeneralization,
  title={Generalization vs Memorization: Tracing Language Models’ Capabilities Back to Pretraining Data},
  author={Wang, Xinyi and Antoniades, Antonis and Elazar, Yanai and Amayuelas, Alfonso and Albalak, Alon and Zhang, Kexun and Wang, William Yang},
  booktitle={The Thirteenth International Conference on Learning Representations}
}

@inproceedings{bengio2009curriculum,
  title={Curriculum learning},
  author={Bengio, Yoshua and Louradour, J{\'e}r{\^o}me and Collobert, Ronan and Weston, Jason},
  booktitle={Proceedings of the 26th annual international conference on machine learning},
  pages={41--48},
  year={2009}
}

@article{soviany2022curriculum,
  title={Curriculum learning: A survey},
  author={Soviany, Petru and Ionescu, Radu Tudor and Rota, Paolo and Sebe, Nicu},
  journal={International Journal of Computer Vision},
  volume={130},
  number={6},
  pages={1526--1565},
  year={2022},
  publisher={Springer}
}

@article{wang2021survey,
  title={A survey on curriculum learning},
  author={Wang, Xin and Chen, Yudong and Zhu, Wenwu},
  journal={IEEE transactions on pattern analysis and machine intelligence},
  volume={44},
  number={9},
  pages={4555--4576},
  year={2021},
  publisher={IEEE}
}

@article{shen2024taskbench,
  title={Taskbench: Benchmarking large language models for task automation},
  author={Shen, Yongliang and Song, Kaitao and Tan, Xu and Zhang, Wenqi and Ren, Kan and Yuan, Siyu and Lu, Weiming and Li, Dongsheng and Zhuang, Yueting},
  journal={Advances in Neural Information Processing Systems},
  volume={37},
  pages={4540--4574},
  year={2024}
}

@inproceedings{chai2023improving,
  title={Improving gradient trade-offs between tasks in multi-task text classification},
  author={Chai, Heyan and Cui, Jinhao and Wang, Ye and Zhang, Min and Fang, Binxing and Liao, Qing},
  booktitle={Proceedings of the 61st Annual Meeting of the Association for Computational Linguistics (Volume 1: Long Papers)},
  pages={2565--2579},
  year={2023}
}

@inproceedings{mathur2020tangled,
  title={Tangled up in BLEU: Reevaluating the Evaluation of Automatic Machine Translation Evaluation Metrics},
  author={Mathur, Nitika and Baldwin, Timothy and Cohn, Trevor},
  booktitle={Proceedings of the 58th Annual Meeting of the Association for Computational Linguistics},
  pages={4984--4997},
  year={2020}
}

@inproceedings{sulem2018bleu,
  title={BLEU is not suitable for the evaluation of text simplification},
  author={Sulem, Elior and Abend, Omri and Rappoport, Ari},
  booktitle={2018 Conference on Empirical Methods in Natural Language Processing, EMNLP 2018},
  pages={738--744},
  year={2018},
  organization={Association for Computational Linguistics}
}

@inproceedings{chitty2024llm,
  title={Llm-inference-bench: Inference benchmarking of large language models on ai accelerators},
  author={Chitty-Venkata, Krishna Teja and Raskar, Siddhisanket and Kale, Bharat and Ferdaus, Farah and Tanikanti, Aditya and Raffenetti, Ken and Taylor, Valerie and Emani, Murali and Vishwanath, Venkatram},
  booktitle={SC24-W: Workshops of the International Conference for High Performance Computing, Networking, Storage and Analysis},
  pages={1362--1379},
  year={2024},
  organization={IEEE}
}

@inproceedings{kwon2023efficient,
  title={Efficient memory management for large language model serving with pagedattention},
  author={Kwon, Woosuk and Li, Zhuohan and Zhuang, Siyuan and Sheng, Ying and Zheng, Lianmin and Yu, Cody Hao and Gonzalez, Joseph and Zhang, Hao and Stoica, Ion},
  booktitle={Proceedings of the 29th symposium on operating systems principles},
  pages={611--626},
  year={2023}
}

@article{yu2020gradient,
  title={Gradient surgery for multi-task learning},
  author={Yu, Tianhe and Kumar, Saurabh and Gupta, Abhishek and Levine, Sergey and Hausman, Karol and Finn, Chelsea},
  journal={Advances in neural information processing systems},
  volume={33},
  pages={5824--5836},
  year={2020}
}

@article{liu2021conflict,
  title={Conflict-averse gradient descent for multi-task learning},
  author={Liu, Bo and Liu, Xingchao and Jin, Xiaojie and Stone, Peter and Liu, Qiang},
  journal={Advances in Neural Information Processing Systems},
  volume={34},
  pages={18878--18890},
  year={2021}
}

@inproceedings{standley2020tasks,
  title={Which tasks should be learned together in multi-task learning?},
  author={Standley, Trevor and Zamir, Amir and Chen, Dawn and Guibas, Leonidas and Malik, Jitendra and Savarese, Silvio},
  booktitle={International conference on machine learning},
  pages={9120--9132},
  year={2020},
  organization={PMLR}
}

@inproceedings{dalvi2018tracking,
  title={Tracking state changes in procedural text: a challenge dataset and models for process paragraph comprehension},
  author={Dalvi, Bhavana and Huang, Lifu and Tandon, Niket and Yih, Wen-tau and Clark, Peter},
  booktitle={Proceedings of the 2018 Conference of the North American Chapter of the Association for Computational Linguistics: Human Language Technologies, Volume 1 (Long Papers)},
  pages={1595--1604},
  year={2018}
}

@article{song2023ego4d,
  title={Ego4d goal-step: Toward hierarchical understanding of procedural activities},
  author={Song, Yale and Byrne, Eugene and Nagarajan, Tushar and Wang, Huiyu and Martin, Miguel and Torresani, Lorenzo},
  journal={Advances in Neural Information Processing Systems},
  volume={36},
  pages={38863--38886},
  year={2023}
}

@inproceedings{kim2023prometheus,
  title={Prometheus: Inducing fine-grained evaluation capability in language models},
  author={Kim, Seungone and Shin, Jamin and Cho, Yejin and Jang, Joel and Longpre, Shayne and Lee, Hwaran and Yun, Sangdoo and Shin, Seongjin and Kim, Sungdong and Thorne, James and others},
  booktitle={The Twelfth International Conference on Learning Representations},
  year={2023}
}

@article{de1995elections,
  title={On elections by ballot},
  author={de Borda, Jean-Charles},
  journal={Classics of social choice, eds. I. McLean, AB Urken, and F. Hewitt},
  pages={83--89},
  year={1995}
}

@inproceedings{siththaranjandistributional,
  title={Distributional Preference Learning: Understanding and Accounting for Hidden Context in RLHF},
  author={Siththaranjan, Anand and Laidlaw, Cassidy and Hadfield-Menell, Dylan},
  booktitle={The Twelfth International Conference on Learning Representations}
}

@inproceedings{ghosh2025onebench,
  title={Onebench to test them all: Sample-level benchmarking over open-ended capabilities},
  author={Ghosh, Adhiraj and Dziadzio, Sebastian and Prabhu, Ameya and Udandarao, Vishaal and Albanie, Samuel and Bethge, Matthias},
  booktitle={Proceedings of the 63rd Annual Meeting of the Association for Computational Linguistics (Volume 1: Long Papers)},
  pages={32445--32481},
  year={2025}
}

@inproceedings{wu2024autogen,
  title={Autogen: Enabling next-gen LLM applications via multi-agent conversations},
  author={Wu, Qingyun and Bansal, Gagan and Zhang, Jieyu and Wu, Yiran and Li, Beibin and Zhu, Erkang and Jiang, Li and Zhang, Xiaoyun and Zhang, Shaokun and Liu, Jiale and others},
  booktitle={First Conference on Language Modeling},
  year={2024}
}

@inproceedings{hong2023metagpt,
  title={MetaGPT: Meta programming for a multi-agent collaborative framework},
  author={Hong, Sirui and Zhuge, Mingchen and Chen, Jonathan and Zheng, Xiawu and Cheng, Yuheng and Wang, Jinlin and Zhang, Ceyao and Wang, Zili and Yau, Steven Ka Shing and Lin, Zijuan and others},
  booktitle={The twelfth international conference on learning representations},
  year={2023}
}

\clearpage

\renewcommand{\abstract}[1]{}
\settopmatter{printccs=false, printfolios=true}

\twocolumn[
  \begin{center}
    {\Huge\sffamily\bfseries Appendix of FM SO.P: A Progressive Task Mixture Framework with Automatic Evaluation for Cross-Domain SOP Understanding}
    \vspace{1.5em}
  \end{center}
]

% \vspace{1em}
% \begin{center}
%     \includegraphics[width=\textwidth]{table3_negative_sample_plot.pdf}
%     \captionof{figure}{\textbf{Negative sample ratio ablation across task mixture stages.} Pass Rate (\%) with varying positive:negative ratios.}
%     % \label{fig:negative_ablation}
% \end{center}
% \vspace{1em}

\section{Stage 1 Prompt}
\begin{tcblisting}{
  listing only,
  listing options={language={}, escapeinside={!!}, breaklines=true, basicstyle=\ttfamily\small},
  breakable
}
You are given a structured dataset containing multiple scenarios across various domains (e.g., banking, DMV, healthcare, hotel, library, online marketplace, university). Each scenario includes:
- initial_database: system state and user account information
- constraint_parameters: system-wide constraints and thresholds
- user_known: information provided by the user
- constraints: logical conditions that must be satisfied
- action_should_succeed: whether the action should succeed (1) or fail (0)
- user_instruction: description of what the user is trying to accomplish
- user_prompt: the actual user request

Your task is to create contrastive question-answer pairs that help a model understand the semantic relationships between constraints, database states, and action outcomes.

For each concept or scenario pattern, generate:

1. A positive QA: a question asking about why an action succeeds/fails given certain conditions, with a "Yes" answer and detailed explanation of the constraint logic.

2. A negative QA: a question presenting a common misconception or incorrect reasoning about why an action succeeds/fails, with a "No" answer and correct explanation.

Focus on:
- Constraint satisfaction logic (e.g., eligibility thresholds, authentication requirements, availability checks)
- Relationships between database state and action outcomes
- Common failure modes and their causes across different domains
- Differences between similar scenarios with different outcomes

Format:

Scenario: [Brief description of the scenario]
Q1: [Question about correct understanding of constraints/outcome]
A1: Yes, [detailed explanation of why the reasoning is correct]
Q2: [Question with incorrect reasoning or common misconception]
A2: No, [correction and proper explanation]

Example (from banking domain):

Scenario: User "john_doe" applies for a credit card with credit_score=750, minimum_credit_score=600, and action_should_succeed=1.

Q1: In this scenario, does the credit card application succeed because the user's credit score (750) meets the minimum required credit score (600)?
A1: Yes, the application succeeds because john_doe's credit score of 750 exceeds the minimum_credit_score constraint of 600. When the user's credit_score in the database meets or exceeds the constraint_parameters minimum threshold, this constraint is satisfied, allowing the action to proceed.

Q2: In this scenario, does the credit card application succeed primarily because the user provided a high monthly income ($5000)?
A2: No, while the monthly income is considered in the application, the success is determined by satisfying the constraint conditions. In this case, the key constraint is the minimum_credit_score check. The credit score of 750 meeting the threshold of 600 is what allows the action to succeed, not just the monthly income value.

Now process the following dataset and generate semantic-level contrastive pairs that help understand domain-specific logic, constraint relationships, and common misconceptions:

{json_data}

Generate 8-12 contrastive pairs covering different aspects such as:
- Authentication/authorization logic (logged_in_user, username existence)
- Threshold-based constraints
- Data validation and user information correctness
- Compound constraints (AND/OR logic in constraint chains)
- Common reasons for action success vs. failure
\end{tcblisting}

\section{Stage 1 Output Examples}
% \begin{tcblisting}{
%   listing only,
%   minted language=markdown,
%   breakable,          % 允许跨页
%   break at=0pt        % 任意位置断开
% }
\begin{tcblisting}{
  listing only,
  listing options={language={}, escapeinside={!!}, breaklines=true, basicstyle=\ttfamily\small},
  breakable
}
{"instruction": "Does the credit card cancellation succeed because the user has a high !\colorbox{yellow}{credit score}!, even though they have an outstanding card balance?", "input": "", "output": "No, having a high credit score doesn't override other constraints. The system uses strict AND logic, meaning all conditions must be satisfied. Even with john_doe's excellent credit score of 750, the cancellation will fail if any other constraint is not met. In this case, the \"no_credit_card_balance_on_card\" constraint requires a zero balance, and if the card has any outstanding balance, the cancellation will fail regardless of credit score. Credit score isn't even part of the cancellation constraints."}
{"instruction": "Does the credit card cancellation succeed because the user has a high !\colorbox{yellow}{credit limit}!, even with a non-zero balance?", "input": "", "output": "No, having a high credit limit doesn't override the zero balance requirement. The constraint explicitly checks that credit_balance equals zero, regardless of the credit_limit. A card with a $10,000 limit and a $1 balance would still fail the cancellation process, while a card with a $100 limit and $0 balance would succeed. The ratio of balance to limit is irrelevant - only the absolute value of the balance matters, and it must be exactly zero."}
{"instruction": "Is the !\colorbox{yellow}{maximum\_owed\_balance}! threshold used to determine if a user can view their credit card information?", "input": "", "output": "No, the maximum_owed_balance threshold is not used for determining access to view credit card information. Jane_doe has an owed_balance of $1000, which exceeds the maximum_owed_balance of $500, but this would not prevent her from viewing her credit card information if she provides correct authentication credentials. This threshold would likely be used for credit applications or increasing credit limits."}
{"instruction": "Does the system require users to have a !\colorbox{yellow}{minimum\_account\_balance\_safety\_box}! amount to access their credit card information?", "input": "", "output": "No, the minimum_account_balance_safety_box parameter (set at $300) is not used as a constraint for accessing credit card information. This parameter likely relates to safety box services, which are separate from credit card information access. The constraints for get_credit_cards only involve authentication factors, not account balance requirements."}
\end{tcblisting}

\section{Stage 2 Prompt}
\begin{tcblisting}{
  listing only,
  listing options={language={}, escapeinside={||}, breaklines=true, basicstyle=\ttfamily\small},
  breakable
}
You are given a structured dataset containing multiple scenarios across various domains (e.g., banking, DMV, healthcare, hotel, library, online marketplace, university). Each scenario includes:
- initial_database: system state and user account information
- constraint_parameters: system-wide constraints and thresholds
- user_known: information provided by the user
- constraints: logical conditions that must be satisfied
- action_should_succeed: whether the action should succeed (1) or fail (0)
- user_instruction: description of what the user is trying to accomplish
- user_prompt: the actual user request

Your task is to create contrastive question-answer pairs that help a model understand the semantic relationships between constraints, database states, and action outcomes.

For each concept or scenario pattern, generate TWO types of contrastive pairs:

**Type A: Constraint Understanding**
1. One positive QA: a question asking about why an action succeeds/fails given certain conditions, with a "Yes" answer and detailed explanation of the constraint logic.

2. Four negative QAs: questions presenting different common misconceptions or incorrect reasoning about why an action succeeds/fails, each with a "No" answer and correct explanation. Each negative should explore a different type of misunderstanding.

**Type B: Workflow and Action Sequencing**
1. Positive workflow QAs: questions about the correct sequence of function calls based on the directed_action_graph, explaining why certain functions must be called before others.

2. Negative workflow QAs: questions presenting incorrect action sequences that violate the directed_action_graph, explaining why they fail even if individual actions succeed.
                   
Focus on:
- Constraint satisfaction logic (e.g., eligibility thresholds, authentication requirements, availability checks)
- Relationships between database state and action outcomes
- Common failure modes and their causes across different domains
- Differences between similar scenarios with different outcomes
- **Action sequencing and workflow dependencies from directed_action_graph**
- **Why certain functions must be called before others (verification before action)**
- **Incorrect sequences that violate the dependency graph**

Format:

**Type A - Constraint Understanding:**
Scenario: [Brief description of the scenario]
Q1: [Question about correct understanding of constraints/outcome]
A1: Yes, [detailed explanation of why the reasoning is correct]
Q2: [Question with incorrect reasoning or common misconception #1]
A2: No, [correction and proper explanation]
Q3: [Question with incorrect reasoning or common misconception #2]
A3: No, [correction and proper explanation]
Q4: [Question with incorrect reasoning or common misconception #3]
A4: No, [correction and proper explanation]
Q5: [Question with incorrect reasoning or common misconception #4]
A5: No, [correction and proper explanation]

**Type B - Workflow Sequencing:**
Q: What is the correct sequence of function calls to [perform action]?
Input: [Relevant parameters from user_known]
A: The correct sequence is:
1. [function_name](...) - [reason]
2. [function_name](...) - [reason]
3. [function_name](...) - [reason]
[Explanation referencing directed_action_graph]

Q: Is this action sequence correct for [performing action]?
Input: [Incorrect sequence]
A: No, this sequence is incorrect because [explain violation of directed_action_graph]. The correct sequence must [explain proper workflow].

Example (from banking domain):

**Type A - Constraint Understanding:**
Scenario: User "john_doe" applies for a credit card with credit_score=750, minimum_credit_score=600, and action_should_succeed=1.

Q1: In this scenario, does the credit card application succeed because the user's credit score (750) meets the minimum required credit score (600)?
A1: Yes, the application succeeds because john_doe's credit score of 750 exceeds the minimum_credit_score constraint of 600. When the user's credit_score in the database meets or exceeds the constraint_parameters minimum threshold, this constraint is satisfied, allowing the action to proceed.

Q2: In this scenario, does the credit card application succeed primarily because the user provided a high monthly income ($5000)?
A2: No, while the monthly income is considered in the application, the success is determined by satisfying the constraint conditions. In this case, the key constraint is the minimum_credit_score check. The credit score of 750 meeting the threshold of 600 is what allows the action to succeed, not just the monthly income value.

Q3: Does the credit card application succeed because the user already has an existing credit card in their account?
A3: No, having an existing credit card does not determine eligibility for a new application. The success is based on satisfying the constraints such as minimum credit score (750 >= 600). The presence of the existing card "[REDACTED:CREDIT_DEBIT_NUMBER]" in john_doe's account is irrelevant to the new application's approval.

Q4: Does the application succeed because the username "john_doe" is alphabetically favorable or commonly used?
A4: No, the username itself has no bearing on the application outcome. The success is determined by constraint validation: the username must exist in the database and the associated account must meet the credit score requirement. Any valid username with a qualifying credit score would succeed.

Q5: Does the application succeed because the total assets ($20000) exceed a specific threshold?
A5: No, there is no total assets threshold in the constraint_parameters. The constraint_parameters only define thresholds for owed_balance, credit_score, account_balance, etc. The total assets value is collected for credit limit calculation purposes, not as a pass/fail criterion for application approval.

**Type B Example - Workflow Sequencing:**

Q6: What is the correct sequence of function calls to apply for a credit card?
Input: username='john_doe', monthly_income=5000, total_assets=20000, identification='padoesshnwojord'
A6: The correct sequence is:
1. login_user(username='john_doe', identification='padoesshnwojord') - Authenticate the user first
2. internal_check_username_exist(username='john_doe') OR internal_get_database() - Verify user exists (satisfies OR constraint)
3. internal_get_credit_score(username='john_doe') - Verify credit eligibility
4. apply_credit_card(username='john_doe', monthly_income=5000, total_assets=20000) - Execute the application
The directed_action_graph shows these dependencies must be satisfied in this order.

Q7: Is this action sequence correct for applying for a credit card?
Input: 1. apply_credit_card(username='john_doe', monthly_income=5000, total_assets=20000)
A7: No, this sequence is incorrect. It violates the directed_action_graph by skipping all prerequisite checks. The graph shows apply_credit_card has AND constraints requiring: (1) user authentication via login_user, (2) username existence verification, and (3) credit score check. Calling apply_credit_card directly will fail to satisfy the workflow constraints even if the function itself executes.

Q8: Why must internal_get_credit_score be called before apply_credit_card even if we already know the user's credit score from the database?
A8: The directed_action_graph explicitly defines this dependency. The apply_credit_card action has an OR node that requires internal_get_credit_score to be called to validate the minimal_eligible_credit_score constraint. Even though the credit score exists in the database, the workflow requires this explicit validation step to ensure the constraint is checked at runtime before the application proceeds.
                   
Now process the following dataset and generate semantic-level contrastive pairs that help understand domain-specific logic, constraint relationships, and common misconceptions:

{json_data}

Generate 6-8 scenario patterns with mixed Type A and Type B QA pairs:
- 4-5 Type A patterns (constraint understanding): each with 1 positive + 4 negative QAs
- 2-3 Type B patterns (workflow sequencing): each with 2-3 positive workflow QAs + 40-60 negative sequence QAs
Total: 140-200 QA pairs covering different aspects such as:
- Authentication/authorization logic (logged_in_user, username existence)
- Threshold-based constraints (credit scores, balances, income requirements)
- Data validation and user information correctness
- Compound constraints (AND/OR logic in constraint chains)
- Common reasons for action success vs. failure
- **Correct function call sequences based on directed_action_graph**
- **Why verification steps must precede action execution**
- **Incorrect sequences that violate workflow dependencies**
- **Difference between action success and workflow constraint satisfaction**
\end{tcblisting}

\section{Stage 2 Output Examples}
\begin{tcblisting}{
  listing only,
  listing options={language={}, escapeinside={||}, breaklines=true, basicstyle=\ttfamily\small},
  breakable
}
{"instruction": "What is the |\colorbox{yellow}{correct sequence}| of function calls to apply for a credit card when only username verification is required?\nInput: username='john_doe', monthly_income=5000, total_assets=20000", "input": "", "output": "The correct sequence is:\n1. internal_check_username_exist(username='john_doe') OR internal_get_database() - Verify the username exists in the system\n2. apply_credit_card(username='john_doe', monthly_income=5000, total_assets=20000) - Process the application\nThis sequence follows the directed_action_graph where apply_credit_card has an OR dependency on either internal_check_username_exist or internal_get_database to verify the username exists before proceeding."}
{"instruction": "What is the |\colorbox{yellow}{minimum sequence}| of function calls required to pay a bill with a credit card?\nInput: username='john_doe', identification='padoesshnwojord', amount=50.0, card_number='2357 1113 1719 2329'", "input": "", "output": "The minimum required sequence is:\n1. login_user(username='john_doe', identification='padoesshnwojord') - Satisfy the logged_in_user constraint\n2. internal_check_username_exist(username='john_doe') OR internal_get_database() - Satisfy the username existence check\n3. pay_bill_with_credit_card(username='john_doe', card_number='2357 1113 1719 2329', amount=50.0) - Execute the payment\nAccording to the directed_action_graph, these steps represent the minimum path that satisfies all the necessary constraints. The not_over_credit_limit check is performed internally during the pay_bill_with_credit_card function execution."}
{"instruction": "What happens in the function call sequence when |\colorbox{yellow}{username verification succeeds but credit}| |\colorbox{yellow}{score verification fails?}|\nInput:** username='john_doe', identification='padoesshnwojord', credit_score=750, minimum_credit_score=800", "input": "", "output": "** The sequence with error handling would be:\n1. login_user(username='john_doe', identification='padoesshnwojord') -> SUCCEEDS\n2. internal_check_username_exist(username='john_doe') -> SUCCEEDS\n3. internal_get_credit_score(username='john_doe') -> RETURNS 750\n4. Credit score verification (750 < 800) -> FAILS\n5. STOP - The workflow terminates with credit score requirement failure\n\nThe system successfully completes authentication and username verification but fails at the credit score check because 750 is below the required 800. According to the directed_action_graph, this verification must succeed before proceeding to set_safety_box. When the verification fails, the workflow terminates with an appropriate error message about insufficient credit score, and the safety box update is not attempted.\n\n**"}
{"instruction": "Is this a |\colorbox{yellow}{valid optimization}| of the loan payment workflow?\nInput: 1. internal_get_database()\n       2. login_user(username='john_doe', identification='padoesshnwojord')\n       3. pay_loan(username='john_doe', pay_owed_amount_request=200.0)", "input": "", "output": "No, this is not a valid optimization. While internal_get_database and login_user are called correctly, the sequence violates the directed_action_graph by skipping the account verification steps. The graph requires:\n1. get_account_balance to verify sufficient funds\n2. get_account_owed_balance to verify the payment amount is valid\nThese steps cannot be skipped even if the database is already retrieved, as they represent explicit constraint checks that must be executed in the workflow. The optimization fails to validate critical payment conditions before executing the transaction."}
\end{tcblisting}

\section{Stage 3 Prompt}
\begin{tcblisting}{
  listing only,
  listing options={language={}, escapeinside={||}, breaklines=true, basicstyle=\ttfamily\small},
  breakable
}
You are given a structured dataset containing multiple scenarios across various domains (e.g., banking, DMV, healthcare, hotel, library, online marketplace, university). Each scenario includes:
- initial_database: system state and user account information
- constraint_parameters: system-wide constraints and thresholds
- user_known: information provided by the user
- constraints: logical conditions that must be satisfied
- action_should_succeed: whether the action should succeed (1) or fail (0)
- user_instruction: description of what the user is trying to accomplish
- user_prompt: the actual user request

Your task is to create contrastive question-answer pairs that help a model understand the semantic relationships between constraints, database states, and action outcomes.

For each concept or scenario pattern, generate:

**Workflow and Action Sequencing - Negative Only**
Generate 64 negative workflow QAs presenting various incorrect action sequences that violate the directed_action_graph, explaining why they fail even if individual actions succeed. Generate diverse error types including: skipped steps, wrong order, missing verification, missing authentication, calling actions without prerequisites, partial sequences, wrong function choices, etc.
                   
Focus on:
- **Incorrect action sequences that violate the directed_action_graph**
- **Why sequences fail when they skip required steps**
- **Why sequences fail when they execute steps in wrong order**
- **Why sequences fail when they miss verification/authentication**
- **Diverse violation patterns of the dependency graph**

Format:

**Workflow Sequencing - Negative Only (64 per scenario):**

Q: In the scenario where [describe specific scenario: user, action, key parameters], given the directed_action_graph with nodes [list key nodes] and connections [list key connections], and the required parameters for [action_name] are [list required parameter names], is this action sequence correct for [performing action]?
Input: [Incorrect sequence - vary the type of error]
A: No, this sequence is incorrect because it violates the directed_action_graph structure. The graph shows [explain specific DAG structure that is violated], but this sequence [explain what the sequence does wrong]. The correct sequence must [explain proper workflow based on DAG].

**Important: Always include the specific scenario context, DAG information, AND required parameters in each question. Focus on workflow violations (missing steps, wrong order, missing required parameters) rather than parameter type mismatches.**

Example (from different domains):

Example 1 - Missing prerequisite steps:
Q1: In the scenario where john_doe wants to apply for a credit card with monthly_income=5000 and total_assets=20000, given the directed_action_graph where node 0 (apply_credit_card) connects to node 1 (AND) which connects to nodes 2 (login_user), 3 (OR), and 6 (internal_get_credit_score), and the expected parameter types for apply_credit_card are (username, monthly_income, total_assets), is this action sequence correct?
Input: 1. apply_credit_card(username='john_doe', monthly_income=5000, total_assets=20000)
A1: No, this sequence is incorrect. The directed_action_graph shows that node 0 (apply_credit_card) has a connection to node 1 (AND operator), meaning all children of the AND node must be executed first. These include node 2 (login_user), node 3 (OR requiring either internal_check_username_exist or internal_get_database), and node 6 (internal_get_credit_score). Calling apply_credit_card directly (node 0) without satisfying nodes 2, 3, and 6 violates the graph structure defined by connection [(0,1), (1,2), (1,3), (1,6)].

Example 2 - Missing required parameters:
Q2: In the scenario where Grimoire wants to add a book to the library with 3 copies, given the directed_action_graph with nodes [add_book, AND, login_user, OR, internal_is_admin, internal_get_database] and connections [(0,1), (1,2), (1,3), (3,4), (3,5)], and the required parameters for add_book are (username, book_title, count, restricted), is this action sequence correct?
Input: 1. login_user(username='Grimoire', password=[REDACTED:PASSWORD])
       2. internal_is_admin(username='Grimoire')
       3. add_book(username='Grimoire', count=3, restricted=False)
A2: No, this sequence is incorrect because it's missing the required parameter 'book_title' in the add_book function call. While the sequence follows the correct workflow order according to the directed_action_graph (login -> verify admin -> add book), the add_book action cannot be completed without specifying which book to add. The user stated they want to add a book, but didn't provide the book_title parameter, making the action impossible to execute meaningfully.        

Now process the following dataset and generate semantic-level contrastive pairs that help understand domain-specific logic, constraint relationships, and common misconceptions:

{json_data}

Generate 64 negative workflow QA pairs per scenario, covering different violation types such as:
- **Skipping authentication steps**
- **Skipping verification/validation steps**
- **Wrong order of function calls**
- **Missing AND constraint satisfaction**
- **Missing OR constraint satisfaction (not calling any option)**
- **Direct action execution without prerequisites**
- **Partial sequences (missing initial steps)**
- **Partial sequences (missing intermediate steps)**
- **Partial sequences (missing final steps)**
- **Using wrong verification/validation functions**
- **Missing required parameters (e.g., trying to add_book without book_title)**
- **Providing wrong entity values (e.g., wrong username, non-existent book)**
- **Mixing up dependency chains**

**Critical: Generate 64 diverse negative examples per scenario. Each negative should demonstrate a different specific way the directed_action_graph can be violated.**

**Focus on workflow violations, not type mismatches: Generate errors related to missing workflow steps, wrong execution order, and missing/wrong required parameter VALUES (e.g., missing book_title when adding a book, wrong username that doesn't exist in database). Do NOT generate errors about parameter types (e.g., 3 vs "three" are semantically equivalent for users).**
\end{tcblisting}

\section{Stage 3 Output Examples}
\begin{tcblisting}{
  listing only,
  listing options={language={}, escapeinside={||}, breaklines=true, basicstyle=\ttfamily\small},
  breakable
}
{"instruction": "In the scenario where john_doe wants to apply for a credit card with monthly_income=5000 and total_assets=20000, given the directed_action_graph where |\colorbox{yellow}{node 0 (apply\_credit\_card) connects to}| |\colorbox{yellow}{node 1 (OR) which connects to nodes 2}| |\colorbox{yellow}{(internal\_check\_username\_exist)}| and 3 (internal_get_database), is this action sequence correct?\nInput: 1. |\colorbox{yellow}{apply\_credit\_card(username='john\_doe',}| |\colorbox{yellow}{monthly\_income=5000, total\_assets=20000)}|", "input": "", "output": "No, this sequence is incorrect because it violates the directed_action_graph structure. The graph shows that node 0 (apply_credit_card) has a connection to node 1 (OR operator), meaning at least one of the children of the OR node must be executed first. These include either node 2 (internal_check_username_exist) or node 3 (internal_get_database). Calling apply_credit_card directly without first satisfying the OR condition by executing either internal_check_username_exist or internal_get_database violates the dependency structure defined by connections [(0,1), (1,2), (1,3)]."}
{"instruction": "In the scenario where john_doe wants to log out of their banking account, given the directed_action_graph where |\colorbox{yellow}{node 0 (logout\_user) connects to node 1}| |\colorbox{yellow}{(OR) which connects to nodes 2}| |\colorbox{yellow}{(internal\_check\_username\_exist) and 3}| |\colorbox{yellow}{(internal\_get\_database)}|, is this action sequence correct for performing a logout?\nInput: 1. internal_check_username_exist(username='|\colorbox{yellow}{jane\_doe}|')       2. logout_user(username='|\colorbox{yellow}{john\_doe}|')", "input": "", "output": "No, this sequence is incorrect because it violates parameter consistency across the directed_action_graph. While the sequence correctly starts with one of the required verification functions (internal_check_username_exist), it checks for username 'jane_doe' but then attempts to log out username 'john_doe'. This inconsistency breaks the logical flow of the graph where the same username parameter should be used throughout the sequence. The correct sequence would use the same username in both function calls to ensure the verified user is the one being logged out."}
{"instruction": "In the scenario where jane_doe wants to set a new admin password to \"newsecurepassword\" using incorrect current admin password, given the directed_action_graph with nodes for |\colorbox{yellow}{set\_admin\_password}|, |\colorbox{yellow}{authenticate\_admin\_password, and}| |\colorbox{yellow}{login\_user, with connections [(0,1),}| |\colorbox{yellow}{(1,2)]}|, is this action sequence correct?**\nInput:\n```\n1. set_admin_password(username='jane_doe', admin_password_new='newsecurepassword')\n2. |\colorbox{yellow}{login\_user}|(username='jane_doe', identification={\"drivers_license_id\": \"D1234567\", \"drivers_license_state\": \"CA\"})\n3. |\colorbox{yellow}{authenticate\_admin\_password}|(username='jane_doe', admin_password='HOjuUuAlomPjs')\n```", "input": "", "output": "** No, this sequence is incorrect because it completely reverses the required order specified in the directed_action_graph. The connections [(0,1), (1,2)] indicate that login_user (node 2) must be executed first, followed by authenticate_admin_password (node 1), and finally set_admin_password (node 0). The provided sequence incorrectly starts with set_admin_password, followed by login_user, and ends with authenticate_admin_password. This violates the logical security flow where authentication must precede any account modifications. Additionally, the admin password \"HOjuUuAlomPjs\" is incorrect, as jane_doe's admin password is \"addoeminnepajass\".\n\n**"}
\end{tcblisting}

\section{Evaluation System Agent 1 Prompt}
\begin{tcblisting}{
  listing only,
  listing options={language={}, escapeinside={||}, breaklines=true, basicstyle=\ttfamily\small},
  breakable
}
You are given a structured dataset containing multiple scenarios across various domains (e.g., banking, DMV, healthcare, hotel, library, online marketplace, university). Each scenario includes:
- initial_database: system state and user account information
- constraint_parameters: system-wide constraints and thresholds
- user_known: information provided by the user
- constraints: logical conditions that must be satisfied
- action_should_succeed: whether the action should succeed (1) or fail (0)
- user_instruction: description of what the user is trying to accomplish
- user_prompt: the actual user request

Your task is to think what aspects are most important to evaluate when an AI answers questions about the scenario.

Now process the following dataset and generate the evaluation criteria with brief descriptions:

{json_data}
\end{tcblisting}

\section{Rubric Example: Bank}
\begin{tcblisting}{
  listing only,
  listing options={language={}, escapeinside={||}, breaklines=true, basicstyle=\ttfamily\small},
  breakable
}
## Authentication Verification
- Properly verifies user credentials (username, password, identification) before proceeding with sensitive actions
- Correctly identifies when multi-factor authentication is required
- Validates admin credentials when necessary for privileged operations
- Rejects transactions when credentials don't match or are invalid

## Security Protocol Adherence
- Follows proper security protocols for different transaction types
- Avoids revealing sensitive information (passwords, full identification details)
- Maintains appropriate information boundaries
- Handles credentials securely without unnecessary repetition

## Constraint Validation
- Correctly identifies and applies all relevant system constraints
- Verifies account existence for all parties in a transaction
- Checks balance sufficiency for withdrawals, payments, and transfers
- Validates that transactions comply with limits (credit limits, maximum transfers)
- Confirms prerequisite conditions are met (minimum balances, credit scores)

## Financial Calculation Accuracy
- Performs precise currency conversions and exchange rate calculations
- Correctly calculates new balances after transactions
- Accurately determines available credit or funds
- Properly handles decimal places and rounding in monetary amounts

## Transaction Outcome Communication
- Clearly states whether a transaction succeeded or failed
- Provides specific reasons for transaction failures
- Confirms details of successful transactions (amounts, new balances)
- Avoids technical jargon in explanations to users

## Error Handling
- Appropriately responds when transactions should fail
- Identifies and communicates specific constraint violations
- Offers constructive guidance when actions cannot be completed
- Suggests viable alternatives when appropriate

## Data Privacy Protection
- Protects sensitive financial and personal information
- Avoids unnecessary exposure of account details
- Does not reveal system parameters or thresholds that could compromise security
- Maintains confidentiality of other users' information

## Response Completeness
- Addresses all aspects of the user's request
- Provides all necessary information about transaction outcomes
- Includes relevant account information when appropriate
- Answers both explicit and implicit questions in user prompts

## User Guidance Quality
- Provides clear next steps when transactions fail
- Explains requirements for successful transactions
- Offers helpful alternatives when primary requests cannot be fulfilled
- Educates users about relevant banking policies when appropriate

## Process Transparency
- Clearly explains what information is needed and why
- Communicates relevant banking policies affecting outcomes
- Provides appropriate context for decisions
- Demonstrates logical reasoning in complex scenarios

## Contextual Appropriateness
- Maintains professional tone suitable for banking context
- Uses relevant financial terminology correctly
- Adapts responses to specific scenario requirements
- Demonstrates awareness of the current system state

## System State Awareness
- Correctly references current account balances and limits
- Recognizes relationships between different account parameters
- Demonstrates understanding of how transactions affect account status
- Applies constraints based on current system conditions
\end{tcblisting}

\section{Evaluation System Agent 2 Prompt}
\begin{tcblisting}{
  listing only,
  listing options={language={}, escapeinside={||}, breaklines=true, basicstyle=\ttfamily\small},
  breakable
}
You are given a structured dataset containing multiple scenarios across various domains (e.g., banking, DMV, healthcare, hotel, library, online marketplace, university). Each scenario includes:
- initial_database: system state and user account information
- constraint_parameters: system-wide constraints and thresholds
- user_known: information provided by the user
- constraints: logical conditions that must be satisfied
- action_should_succeed: whether the action should succeed (1) or fail (0)
- user_instruction: description of what the user is trying to accomplish
- user_prompt: the actual user request

Your task is to generate difficult or nuanced questions a user might ask about these SOPs, and provide the correct answer based on the contents.

Format:

## Question 1: xxx

**Anwser:** xxx

## Question 2: xxx

**Anwser:** xxx

......

## Question N: xxx

**Anwser:** xxx

Now process the following dataset and generate the questions and answers:

{json_data}
\end{tcblisting}

\section{Test Set Example: Bank}
\begin{tcblisting}{
  listing only,
  listing options={language={}, escapeinside={||}, breaklines=true, basicstyle=\ttfamily\small},
  breakable
}
{"instruction": "How does the system determine the credit limit for a new credit card?", "input": "", "output": "The system appears to determine the credit limit based on a combination of factors including the user's monthly income, total assets, credit score, and possibly their existing account balance. While the exact formula isn't explicitly stated, successful applications show that users with higher incomes, assets, and credit scores are likely to receive higher credit limits."}
{"instruction": "What information is stored about a user's account in the bank's database?", "input": "", "output": "The bank's database stores comprehensive account information including: identification code, admin password, account balance, owed balance, credit score, safety box contents, and credit card details (including card number, credit limit, and credit balance)."}
{"instruction": "What currencies are currently available for exchange at the bank?", "input": "", "output": "Based on the database, the bank currently offers exchange services for four foreign currencies: Euro (EUR), Chinese Yuan/Renminbi (RMB), British Pound (GBP), and New Taiwan Dollar (NTD). The current exchange rates are 0.93 EUR, 7.12 RMB, 0.77 GBP, and 32.08 NTD per US dollar."}
{"instruction": "How is the amount of foreign currency I receive calculated?", "input": "", "output": "The amount of foreign currency you receive is calculated by multiplying your USD amount by the exchange rate for the requested currency. For example, if you exchange $100 USD for EUR at an exchange rate of 0.93, you would receive 93 EUR (100 x 0.93 = 93)."}
\end{tcblisting}

\section{Evaluation System Agent 3 Prompt}
\begin{tcblisting}{
  listing only,
  listing options={language={}, escapeinside={||}, breaklines=true, basicstyle=\ttfamily\small},
  breakable
}
Question: {question}
Ground truth Answer: {gt_answer}
Model Answer: {model_answer}

Evaluate the Model's answer based on the follwoing criteria. Please make sure **EACH ASPECT** (start with ##) is evaluated.

=== criteria begins ===

{ruberic}

=== criteria ends ===

Provide a score 0 - 10 (or judgement) for each aspect with reasoning. Format:

## [Aspect 1 name]: x/10

[reason]

## [Aspect 2 name]: x/10

[reason]

......

## Overall Assessment:

......
\end{tcblisting}

\section{Scoring Example: Bank}
\begin{tcblisting}{
  listing only,
  listing options={language={}, escapeinside={||}, breaklines=true, basicstyle=\ttfamily\small},
  breakable
}
{
  "question_id": 11,
  "scores": {
    "authentication_verification": "8/10",
    "security_protocol_adherence": "9/10",
    "constraint_validation": "7/10",
    "financial_calculation_accuracy": "N/A",
    "transaction_outcome_communication": "9/10",
    "error_handling": "8/10",
    "data_privacy_protection": "10/10",
    "response_completeness": "7/10",
    "user_guidance_quality": "8/10",
    "process_transparency": "9/10",
    "contextual_appropriateness": "10/10",
    "system_state_awareness": "8/10"
  ,
  "result": "
# Evaluation of model's Answer on Credit Card Cancellation Requirements

## Authentication Verification: 8/10
The model correctly identifies that proper authentication is required (logged_in_user), which aligns with the ground truth. However, it doesn't explicitly mention that either login credentials or admin password would work, which is a minor omission.

## Security Protocol Adherence: 9/10
The answer maintains appropriate information boundaries and doesn't reveal any sensitive information. It properly focuses on the procedural requirements without exposing system vulnerabilities.

## Constraint Validation: 7/10
The model correctly identifies the zero balance requirement but adds an incorrect constraint about "zero credit limit" that isn't mentioned in the ground truth. It correctly notes the card must exist in the user's account, but doesn't explicitly state that the username must exist in the system (though this may be implied by authentication).

## Financial Calculation Accuracy: N/A
This criterion isn't applicable to this particular question as no financial calculations are involved.

## Transaction Outcome Communication: 9/10
The answer clearly states that all conditions must be met for cancellation to succeed and that failure to meet any requirement will result in cancellation failure, which is accurate and helpful.

## Error Handling: 8/10
The model appropriately explains that the action will fail if requirements aren't met, providing clear guidance on potential failure points.

## Data Privacy Protection: 10/10
The answer doesn't reveal any sensitive information or system parameters that could compromise security.

## Response Completeness: 7/10
The answer covers most key points but includes an incorrect requirement (zero credit limit) while missing the explicit mention that the username must exist in the system.

## User Guidance Quality: 8/10
The answer clearly explains requirements for successful cancellation, which provides good guidance to users wanting to cancel their credit cards.

## Process Transparency: 9/10
The model provides clear explanation of what's needed and why (preventing cancellation of active cards with balances), offering good context for the requirements.

## Contextual Appropriateness: 10/10
The answer maintains a professional tone appropriate for banking context and uses relevant terminology correctly.

## System State Awareness: 8/10
The model demonstrates good understanding of how the system checks for card existence and balance before allowing cancellation.

## Overall Assessment:
The model's answer is mostly accurate and comprehensive, scoring well across most criteria. The main issue is the inclusion of an incorrect requirement (zero credit limit) that isn't mentioned in the ground truth. It also doesn't explicitly state that the username must exist in the system, though this might be considered implicit in the authentication requirement. The answer provides clear guidance on what's needed for successful cancellation and explains why these requirements exist, which is helpful for users. Overall, the response is professional, secure, and mostly accurate despite the noted discrepancies."
}
{
  "question_id": 23,
  "scores": {
    "authentication_verification": "2/10",
    "security_protocol_adherence": "6/10",
    "constraint_validation": "3/10",
    "financial_calculation_accuracy": "5/10",
    "transaction_outcome_communication": "4/10",
    "error_handling": "3/10",
    "data_privacy_protection": "7/10",
    "response_completeness": "4/10",
    "user_guidance_quality": "3/10",
    "process_transparency": "4/10",
    "contextual_appropriateness": "6/10",
    "system_state_awareness": "2/10"
  ,
  "result": "
# Evaluation of model's Answer

## Authentication Verification: 2/10

The model completely misidentifies the authentication issue. The ground truth clearly states that the failure was due to an incorrect admin password ("fjtjpjpeUCLbMGlOCQq" vs. the correct "addoeminhnpajoss"), but the model incorrectly claims the identification was correct and attributes the failure to a credit card balance constraint instead.

## Security Protocol Adherence: 6/10

The model doesn't reveal sensitive information like passwords, which is good. However, it fundamentally misunderstands the security protocol that was violated, focusing on constraints rather than authentication requirements.

## Constraint Validation: 3/10

The model invents a constraint ("no_credit_card_balance_on_card") that isn't mentioned in the ground truth. While checking for zero balances before account closure might be a reasonable constraint in banking systems, the answer fabricates this as the reason for failure when the actual issue was authentication.

## Financial Calculation Accuracy: 5/10

The model mentions specific financial details ($200 balance on a credit card) that aren't referenced in the ground truth. While the amount itself isn't calculated incorrectly, the relevance of this information to the account closure failure is fabricated.

## Transaction Outcome Communication: 4/10

The model clearly states why it believes the transaction failed, but the explanation is incorrect. It provides a specific but fabricated reason for the failure.

## Error Handling: 3/10

The model identifies what it believes is the constraint violation, but since this is not the actual reason for failure, the error handling is fundamentally flawed.

## Data Privacy Protection: 7/10

The model doesn't reveal sensitive authentication details, which is appropriate. It does mention a specific credit card number, but formats it in the standard masked way.

## Response Completeness: 4/10

The answer addresses the question but with incorrect information. It provides a complete but inaccurate explanation of why the account closure failed.

## User Guidance Quality: 3/10

The model doesn't provide guidance on how to successfully close the account. Since it misidentifies the problem as a credit card balance issue rather than an authentication issue, any implied guidance would lead the user in the wrong direction.

## Process Transparency: 4/10

The model attempts to explain the process logically but bases its explanation on incorrect assumptions about what constraints were violated.

## Contextual Appropriateness: 6/10

The tone is professional and appropriate for a banking context, and the model uses relevant financial terminology correctly. However, it demonstrates a fundamental misunderstanding of the scenario.

## System State Awareness: 2/10

The model shows poor awareness of the actual system state. It invents details about credit card balances and constraints that aren't mentioned in the ground truth, while missing the actual authentication failure that occurred.

## Overall Assessment:

The model's answer is fundamentally flawed because it completely misidentifies the reason for the account closure failure. The ground truth clearly states that the failure was due to an incorrect admin password, but the model fabricates an entirely different reason involving credit card balances. This demonstrates a critical failure in understanding the authentication requirements for account closure. While the answer is presented professionally and avoids revealing sensitive information, its core explanation is incorrect, which would mislead users about how to successfully complete their intended action. The invention of specific details not present in the ground truth (credit card numbers, balances, and constraints) is particularly concerning as it shows the model is confidently providing fabricated information."
}
\end{tcblisting}

\section{Scoring Example: DMV}
\begin{tcblisting}{
  listing only,
  listing options={language={}, escapeinside={||}, breaklines=true, basicstyle=\ttfamily\small},
  breakable
}
{
  "question_id": 11,
  "scores": {
    "authentication_verification": "5/10",
    "data_validation": "6/10",
    "constraint_compliance": "3/10",
    "temporal_reasoning": "4/10",
    "error_handling": "0/10",
    "privacy_protection": "7/10",
    "success_confirmation": "0/10",
    "system_state_awareness": "4/10",
    "process_transparency": "6/10",
    "next_steps_guidance": "0/10",
    "eligibility_assessment": "5/10",
    "resource_availability_checking": "3/10"
  ,
  "result": "
# Evaluation of model's Answer on Driver's License Address Change Requirements

## Authentication Verification: 5/10

The model mentions "user must be logged in with valid credentials" and "user must have a valid identification document," which acknowledges authentication requirements. However, it misses the specific ground truth requirement of authenticating with identification/password in some cases. The answer is too generic and doesn't capture the system's actual authentication process.

## Data Validation: 6/10

The model correctly notes that the license must exist in the system and that updated address information must be provided. However, it fails to mention the critical requirement that the new address must be different from the current one, which is explicitly stated in the ground truth.

## Constraint Compliance: 3/10

The model introduces many constraints that aren't mentioned in the ground truth, such as "license must not be suspended," "license must not be revoked," "license must be in good standing," etc. These fabricated requirements mislead users about the actual system constraints.

## Temporal Reasoning: 4/10

The model incorrectly introduces a temporal constraint about the license not being expired ("must have an active driver's license"), which isn't mentioned in the ground truth. This represents an incorrect understanding of the system's temporal requirements.

## Error Handling: 0/10

The answer doesn't address error handling at all. It doesn't explain what happens if requirements aren't met or what errors might be encountered.

## Privacy Protection: 7/10

The answer appropriately acknowledges the need for authentication and verification without exposing sensitive details about the process.

## Success Confirmation: 0/10

The answer doesn't mention what happens after requirements are met or how success is confirmed to the user.

## System State Awareness: 4/10

The model shows partial awareness of the system state by mentioning the license must exist, but it introduces many incorrect states like "suspended," "revoked," and "license_restricted" that aren't mentioned in the ground truth.

## Process Transparency: 6/10

The answer provides a clear list of requirements, making the process somewhat transparent, but many of these requirements are fabricated and not part of the actual system.

## Next Steps Guidance: 0/10

The answer doesn't provide any guidance on next steps or what happens after the requirements are met.

## Eligibility Assessment: 5/10

The model correctly identifies that a valid license is required but adds many incorrect eligibility criteria not present in the ground truth.

## Resource Availability Checking: 3/10

The answer mentions checking if the license exists but doesn't address other resource availability aspects correctly.

## Overall Assessment:

The model's answer is significantly flawed and potentially misleading. While it correctly identifies some basic requirements (having a valid license, providing updated address information), it fabricates numerous requirements not present in the ground truth, such as license suspension status, outstanding violations, and state jurisdiction limitations. It completely misses key actual requirements like the new address needing to be different from the current one. The answer demonstrates a tendency to generate plausible-sounding but incorrect information rather than adhering to the actual system requirements. This could lead users to believe they need to meet many unnecessary conditions to change their address, potentially causing confusion and frustration."
}
{
  "question_id": 42,
  "scores": {
    "authentication_verification": "4/10",
    "data_validation": "2/10",
    "constraint_compliance": "3/10",
    "temporal_reasoning": "1/10",
    "error_handling": "0/10",
    "privacy_protection": "3/10",
    "success_confirmation": "0/10",
    "system_state_awareness": "2/10",
    "process_transparency": "3/10",
    "next_steps_guidance": "0/10",
    "eligibility_assessment": "1/10",
    "resource_availability_checking": "0/10"
  ,
  "result": "
# Evaluation of model's Answer on Driver's License Renewal Authentication Methods

## Authentication Verification: 4/10

The model's answer incorrectly identifies three authentication methods (login_user, login_driver, authenticate_admin_password) that don't match the ground truth. The ground truth specifies only two methods: simple authentication (username only) and full authentication (username + identification number). The model appears to be listing API functions rather than explaining the actual authentication approaches required for license renewal.

## Data Validation: 2/10

The answer doesn't address how the system validates authentication data, particularly the critical point from the ground truth that identification numbers must match exactly what's in the system. The example of an incorrect identification number causing renewal failure is completely missing.

## Constraint Compliance: 3/10

The model fails to mention the important constraint that simple authentication only works when renewal conditions are met and the user is within the renewal window. These business rules are central to understanding the authentication requirements.

## Temporal Reasoning: 1/10

The answer completely omits the temporal aspect mentioned in the ground truth - that simple authentication only works within the renewal window. This is a critical omission as it relates to when different authentication methods are required.

## Error Handling: 0/10

The model provides no information about what happens when authentication fails, unlike the ground truth which explicitly states that renewal will fail if an incorrect identification number is provided.

## Privacy Protection: 3/10

While the answer mentions different authentication scopes, it doesn't explain the security implications of the different methods as described in the ground truth, which notes that full authentication provides "more security."

## Success Confirmation: 0/10

The answer doesn't address what happens when authentication succeeds or fails.

## System State Awareness: 2/10

The model mentions a "directed_action_graph" that isn't referenced in the ground truth, suggesting it's pulling information from elsewhere rather than addressing the specific question about authentication methods for license renewal.

## Process Transparency: 3/10

The answer attempts to explain different authentication methods but misses the actual processes described in the ground truth. It doesn't clearly explain when each method is used in the renewal process.

## Next Steps Guidance: 0/10

The answer provides no guidance on what users should do if authentication fails or which method to use in different scenarios.

## Eligibility Assessment: 1/10

The model fails to connect authentication methods to eligibility for renewal, which is a key point in the ground truth (simple authentication works only if renewal conditions are met).

## Resource Availability Checking: 0/10

Not applicable to this question about authentication methods.

## Overall Assessment:

The model's answer is fundamentally incorrect and appears to be fabricating authentication methods that don't match the ground truth. It describes API functions rather than the actual authentication approaches for license renewal. The answer misses critical information about when different authentication methods are required, what happens when authentication fails, and the security implications of each method. The model seems to be drawing from system implementation details rather than addressing the user-facing authentication requirements described in the ground truth. Overall, this answer would mislead users about how to authenticate for license renewal."
}
\end{tcblisting}

\begin{figure*}[t!]
    \centering
    \includegraphics[width=\textwidth]{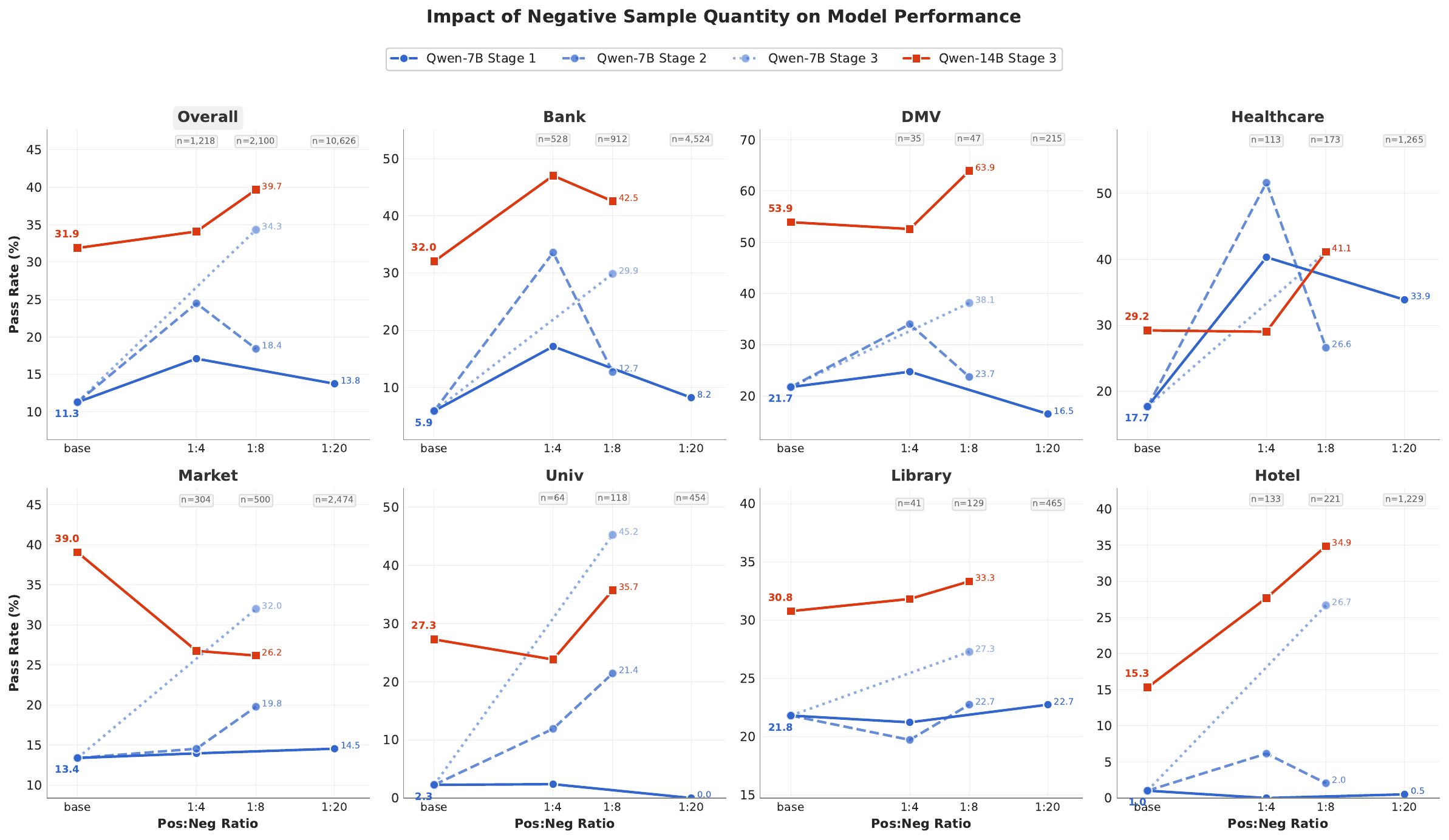}
    \caption{\textbf{Negative sample ratio ablation across task mixture stages.} Pass Rate (\%) with varying positive:negative ratios.}
    % \label{fig:negative_ablation}
\end{figure*}

\section{Domain Analysis of Negative Sample Ratio}
Domain-specific analysis reveals how different task types benefit from varying negative ratios. Banking domain shows Stage 1 highest at 1:4 (17.2\%), with significant drops at 1:20 (8.2\%). Healthcare shows larger Stage 1 improvement, jumping from baseline (17.7\%) to 1:4 (40.3\%), then declining at 1:20 (33.9\%). This U-shaped degradation pattern confirms that sequential reasoning suffers when presented with too many procedural error patterns. DMV demonstrates Stage 2's sensitivity to negative ratio, with 1:4 achieving strong performance (34.0\%) compared to lower performance at base (21.7\%) and 1:8 ratios (23.7\%). University domain validates Stage 3's requirement for higher negative sampling. The performance shows steady improvement from baseline (2.3\%) to highest at 1:8 (45.2\%). Market domain also achieves highest performance (32.0\%) at 1:8. This pattern confirms that graph constraint reasoning requiring enough negative coverage. Library maintaining relatively stable performance across ratios (21.8-27.3\%) and Hotel showing clear 1:8 highest (26.7\% vs 1.0-6.1\% at other ratios). 

% ============================================================================
% Case Study 1: Healthcare Domain - Insurance Claim Appeal Processing
% ============================================================================

\section{Case Study 1: Healthcare Domain}

\subsection{Scenario}
A patient seeks to appeal a denied insurance claim through the healthcare provider's system.

\subsection{Generated SOP}

\paragraph{Step 1: User Authentication}
\textbf{Action:} Verify user identity

\begin{lstlisting}[language=Python, basicstyle=\ttfamily\small]
CALL: login_user(username, identification)
\end{lstlisting}

\textbf{Decision Point 1:} Is user authenticated?
\begin{itemize}
    \item \textbf{IF NO} $\rightarrow$ \textbf{TERMINATE}: Return error ``Authentication Failed''
    \item \textbf{IF YES} $\rightarrow$ \textbf{PROCEED} to Step 2
\end{itemize}

\paragraph{Step 2: Policy Verification}
\textbf{Action:} Validate policy status

\begin{lstlisting}[language=Python, basicstyle=\ttfamily\small]
CALL: get_policy_details(username)
AND: policy_active(username)
\end{lstlisting}

\textbf{Decision Point 2:} Is policy currently active?
\begin{itemize}
    \item \textbf{IF NO} $\rightarrow$ \textbf{TERMINATE}: Return error ``Policy Inactive''
    \item \textbf{IF YES} $\rightarrow$ \textbf{PROCEED} to Step 3
\end{itemize}

\paragraph{Step 3: Claim Existence Verification}
\textbf{Action:} Confirm claim exists in user's history

\begin{lstlisting}[language=Python, basicstyle=\ttfamily\small]
CALL: internal_check_claim_exists(username, claim_id)
\end{lstlisting}

\textbf{Decision Point 3:} Does claim exist AND is status = DENIED?
\begin{itemize}
    \item \textbf{IF NO} $\rightarrow$ \textbf{TERMINATE}: Return error ``Claim Not Found or Not Denied''
    \item \textbf{IF YES} $\rightarrow$ \textbf{PROCEED} to Step 4
\end{itemize}

\paragraph{Step 4: Appeal Period Verification}
\textbf{Action:} Verify appeal is within allowable timeframe

\begin{lstlisting}[language=Python, basicstyle=\ttfamily\small]
CALL: within_appeal_timeframe(claim_id, current_date)
RULE: appeal_date - denial_date <= 30 days
\end{lstlisting}

\textbf{Decision Point 4:} Is appeal within 30-day window?
\begin{itemize}
    \item \textbf{IF NO} $\rightarrow$ \textbf{TERMINATE}: Return error ``Appeal Period Expired''
    \item \textbf{IF YES} $\rightarrow$ \textbf{PROCEED} to Step 5
\end{itemize}

\paragraph{Step 5: Appeal Processing}
\textbf{Action:} Create appeal record and initiate review

\begin{lstlisting}[language=Python, basicstyle=\ttfamily\small]
EXECUTE:
  - Generate appeal_id
  - Set status = "PENDING_REVIEW"
  - Assign to review queue
  - Send confirmation to user
\end{lstlisting}

\textbf{Outcome:} \textsc{Success} - Appeal submitted for review

\subsection{Key Observations}
This SOP demonstrates several features. First, it implements sequential validation where four decision points must pass in order. Second, it includes early termination logic where any failed check immediately terminates the workflow. Third, it enforces temporal constraints through appeal timeframe validation to ensure policy compliance. Finally, it performs state verification by validating claim status before proceeding with subsequent operations.

% ============================================================================
% Case Study 2: Hotel Domain - Loyalty Program Registration
% ============================================================================

\section{Case Study 2: Hotel Domain}

\subsection{Scenario}
A guest staying at the hotel wishes to join the loyalty rewards program.

\subsection{Generated SOP}

\paragraph{Step 1: Registration Initiation}
\textbf{Action:} Begin loyalty registration process

\begin{lstlisting}[language=Python, basicstyle=\ttfamily\small]
CALL: register_loyalty_member(guest_name="Jane Smith")
\end{lstlisting}

\paragraph{Step 2: Membership Status Check}
\textbf{Action:} Query existing membership database

\begin{lstlisting}[language=Python, basicstyle=\ttfamily\small]
CALL: internal_is_loyalty_member(guest_name="Jane Smith")
\end{lstlisting}

\textbf{Decision Point 1:} Is guest already enrolled?
\begin{itemize}
    \item \textbf{IF YES} $\rightarrow$ \textbf{TERMINATE}: Return message ``Already a loyalty member''
    \item \textbf{IF NO} $\rightarrow$ \textbf{PROCEED} to Step 3
\end{itemize}

\paragraph{Step 3: Room Eligibility Verification}
\textbf{Action:} Verify room type meets program requirements

\begin{lstlisting}[language=Python, basicstyle=\ttfamily\small]
CHECK: current_reservation.room_type
ELIGIBLE: Single Room, Double Room, Suite
\end{lstlisting}

\textbf{Decision Point 2:} Is room type eligible?
\begin{itemize}
    \item \textbf{IF NO} $\rightarrow$ \textbf{TERMINATE}: Return error ``Room Not Eligible''
    \item \textbf{IF YES} $\rightarrow$ \textbf{PROCEED} to Step 4
\end{itemize}

\paragraph{Step 4: Member Profile Creation}
\textbf{Action:} Create new member record

\begin{lstlisting}[language=Python, basicstyle=\ttfamily\small]
EXECUTE:
  - Add guest to loyalty_members database
  - Generate unique member_id = "HTL2007Z"
  - Set initial tier = "Silver"
  - Initialize points balance = 0
  - Set enrollment_date = current_date
\end{lstlisting}

\textbf{Outcome:} \textsc{Success} - Member registered with ID ``HTL2007Z''

\subsection{Key Observations}
This SOP demonstrates several features. First, it implements duplicate prevention by checking existing membership before registration. Second, it enforces eligibility criteria where room type determines program qualification. Third, it performs state initialization by assigning new members a default tier and zero points. Finally, it uses nested decisions through two sequential checks before profile creation.

% ============================================================================
% Case Study 3: E-commerce Domain - Product Review Submission
% ============================================================================

\subsection{Case Study 3: E-commerce Domain}

\subsubsection{Scenario}
A customer wants to submit a review for a purchased product.

\subsubsection{Generated SOP}

\paragraph{Step 1: User Authentication}
\textbf{Action:} Verify user identity (OR logic - multiple methods accepted)

\begin{lstlisting}[language=Python, basicstyle=\ttfamily\small]
METHOD 1: login_user(username, password)
   OR
METHOD 2: internal_check_username_exist(username)
\end{lstlisting}

\paragraph{Step 2: Eligibility Verification}
\textbf{Action:} Verify ALL eligibility requirements (AND logic)

\textbf{Critical:} ALL five conditions must be satisfied simultaneously.

\textbf{Condition 1 - Rating Validity:}
\begin{lstlisting}[language=Python, basicstyle=\ttfamily\small]
CALL: within_review_limits(rating)
REQUIREMENT: 1 <= rating <= 5
\end{lstlisting}

\textbf{Condition 2 - Duplicate Prevention:}
\begin{lstlisting}[language=Python, basicstyle=\ttfamily\small]
CALL: unique_review(username, product_id)
REQUIREMENT: No existing review by this user for this product
\end{lstlisting}

\textbf{Condition 3 - Account Status:}
\begin{lstlisting}[language=Python, basicstyle=\ttfamily\small]
CALL: credit_status_not_restricted_or_suspended(username)
REQUIREMENT: Account status in {"active", "verified"}
\end{lstlisting}

\textbf{Condition 4 - Purchase Verification:}
\begin{lstlisting}[language=Python, basicstyle=\ttfamily\small]
CALL: product_bought_by_user(username, product_id)
REQUIREMENT: Verified purchase in order history
\end{lstlisting}

\textbf{Condition 5 - Product Existence:}
\begin{lstlisting}[language=Python, basicstyle=\ttfamily\small]
CALL: internal_check_product_exist(product_id)
REQUIREMENT: Product exists in catalog
\end{lstlisting}

\textbf{Decision Point:} Are ALL five conditions satisfied?
\begin{itemize}
    \item \textbf{IF ANY FAILS} $\rightarrow$ \textbf{TERMINATE}: Return specific error message
    \item \textbf{IF ALL PASS} $\rightarrow$ \textbf{PROCEED} to Step 3
\end{itemize}

\paragraph{Step 3: Review Submission}
\textbf{Action:} Create review record

\begin{lstlisting}[language=Python, basicstyle=\ttfamily\small]
CALL: add_review(username, product_id, rating, comment)
CREATE RECORD:
  - review_id: auto-generated UUID
  - verified_purchase: TRUE
  - status: "PENDING_MODERATION"
  - submission_date: current_timestamp
\end{lstlisting}

\textbf{Outcome:} \textsc{Success} - Review submitted and queued for publication

\subsection{Key Observations}
This SOP demonstrates several features. First, it implements OR logic for authentication, providing flexible authentication paths for users. Second, it enforces AND logic for validation where all five conditions must be satisfied simultaneously. Third, it includes anti-fraud measures through purchase verification and duplicate detection mechanisms. Finally, it performs multi-constraint validation to enforce complex business rules systematically.

\end{document}